\documentclass[10pt,twocolumn,letterpaper]{article}

\usepackage[table]{xcolor}
\usepackage{booktabs,graphicx,multirow,wrapfig,caption,subcaption,colortbl,adjustbox,subcaption}
\usepackage[accsupp]{axessibility}

\usepackage[pagenumbers]{cvpr} 

\newcommand{\TODO}[1]{\textbf{\color{red}[TODO: #1]}}
\renewcommand{\TODO}[1]{}

\definecolor{cvprblue}{rgb}{0.21,0.49,0.74}
\usepackage[pagebackref,breaklinks,colorlinks,allcolors=cvprblue]{hyperref}

\def\paperID{34416} 
\def\confName{CVPR}
\def\confYear{2026}

\title{Sparsity as a Key: Unlocking New Insights from Latent Structures\\for Out-of-Distribution Detection}

\author{
Ahyoung Oh$^\dagger$ \quad Wonseok Shin$^\dagger$ \quad Songkuk Kim$^\ddagger$
\\
School of Integrated Technology, BK21 Graduate Program in Intelligent Semiconductor Technology\\
Yonsei University\\
{\tt\small \{ahyoung.oh, wonseok.shin, songkuk\}@yonsei.ac.kr}
}
\graphicspath{{figures/}{sec/}}

\begin{document}
\maketitle

\renewcommand{\thefootnote}{\fnsymbol{footnote}}
\footnotetext[2]{Equal contribution.}
\footnotetext[3]{Corresponding author.}
\renewcommand{\thefootnote}{\arabic{footnote}}

\begin{abstract}
Sparse Autoencoders (SAEs) have demonstrated significant success in interpreting Large Language Models (LLMs) by decomposing dense representations into sparse, semantic components. However, their potential for analyzing Vision Transformers (ViTs) remains largely under-explored. In this work, we present the first application of SAEs to the ViT \texttt{[CLS]} token for out-of-distribution (OOD) detection, addressing the limitation of existing methods that rely on entangled feature representations. We propose a novel framework utilizing a Top-$k$ SAE to disentangle the dense \texttt{[CLS]} features into a structured latent space. Through this analysis, we reveal that in-distribution (ID) data exhibits consistent, class-specific activation patterns, which we formalize as \textbf{Class Activation Profiles (CAPs)}. Our study uncovers a key structural invariant: while ID samples preserve a stable pattern within CAPs, OOD samples systematically disrupt this structure. Leveraging this insight, we introduce a scoring function based on the divergence of core energy profiles to quantify the deviation from ideal activation profiles. Our method achieves strong results on the FPR95 metric—critical for safety-sensitive applications—across multiple benchmarks, while also achieving competitive AUROC. Overall, our findings demonstrate that the sparse, disentangled features revealed by SAEs can serve as a powerful, interpretable tool for robust OOD detection in vision models.
\end{abstract}

\section{Introduction}
\label{sec:intro}

\begin{figure*}[t]
  \centering
   \includegraphics[width=0.98\linewidth]{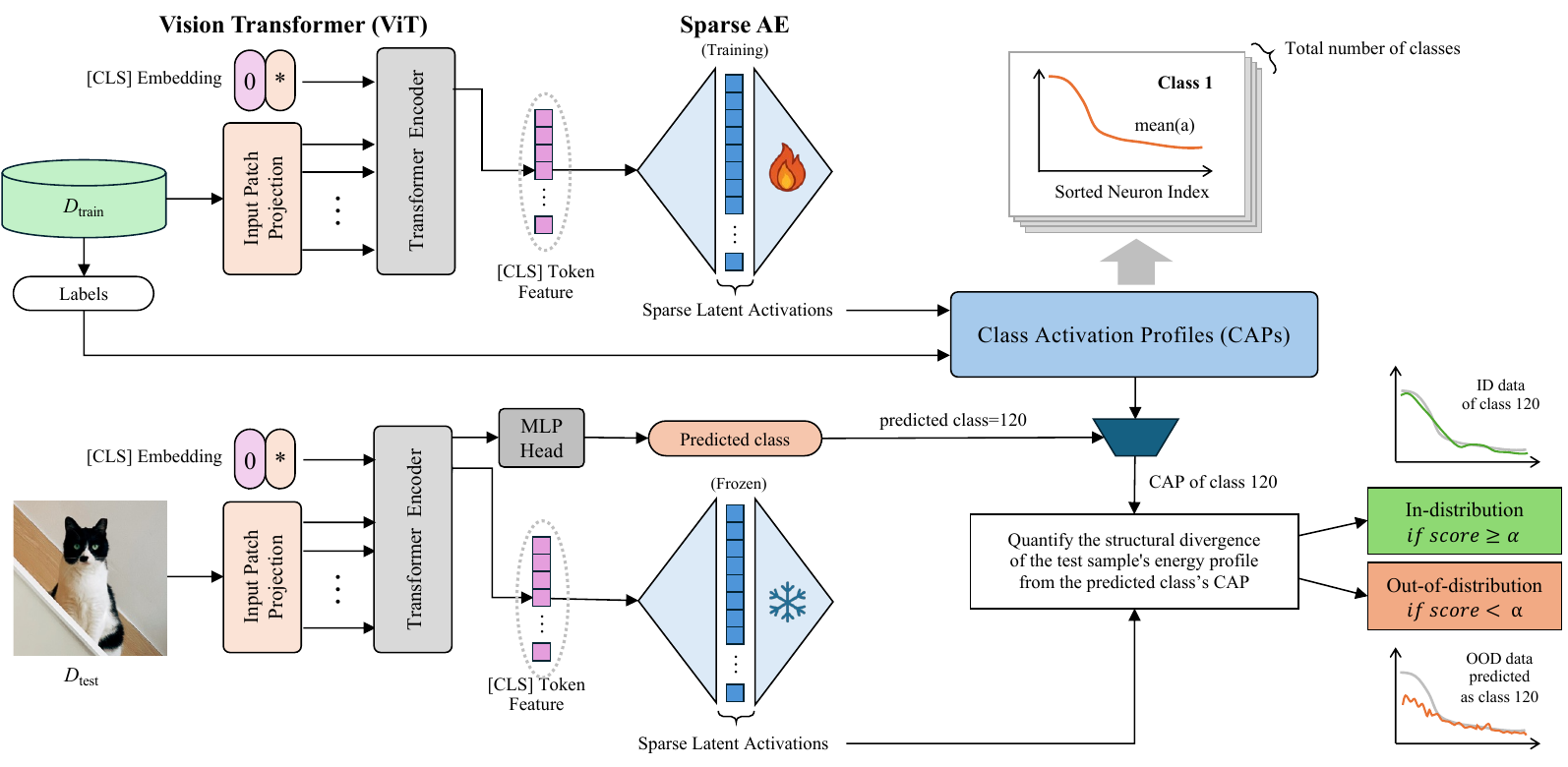}
   \caption{\textbf{Overview of our OOD detection framework.} In the (a) Setup Phase, a Top-$k$ SAE is trained on \texttt{[CLS]} tokens extracted from a fixed, pre-trained ViT using ID data. This process learns disentangled latent features, which are aggregated to form CAPs. In the (b) Inference Phase, the latent activation distribution of a test sample is compared against the predicted class's CAP using our Energy Profile Divergence (EPD) score to compute the final OOD score.}
   \label{fig1:overview}
\end{figure*}

The Vision Transformer (ViT) has achieved remarkable success across a wide range of computer vision tasks~\cite{dosovitskiy2020image}. This success, however, largely depends on the assumption that inference data is drawn from the same distribution as the training set, referred to as in-distribution (ID) data. In real-world scenarios such as autonomous driving or medical diagnosis, this assumption is frequently violated~\cite{hendrycks2021unsolved}. Models inevitably encounter out-of-distribution (OOD) inputs that fall outside their training domain~\cite{hendrycks2016baseline}. Detecting such OOD samples is therefore essential for developing trustworthy AI systems~\cite{fort2021exploring, yang2024generalized}.

Over the past years, numerous methods have been proposed to address this challenge~\cite{yang2024generalized}. Baseline approaches like Maximum Softmax Probability (MSP)~\cite{hendrycks2016baseline} and ODIN~\cite{liang2017enhancing} utilize logit statistics, while energy-based scores~\cite{liu2020energy} refine this by aggregating logit information. Other feature-based methods leverage distance metrics, such as the Mahalanobis distance~\cite{lee2018simple} or non-parametric $k$-nearest neighbors (KNN)~\cite{sun2022out}. Despite their utility, these approaches treat feature vectors as monolithic and opaque. By relying on aggregate statistics—such as magnitude or Euclidean distance—they overlook the fact that deep representations are inherently entangled, with distinct semantic concepts superposed within dense activations~\cite{elhage2022toy}. Consequently, distance-based methods often fail to distinguish spurious geometric proximity from genuine semantic similarity, focusing on the quantity of activation rather than the quality of the internal structure.

This challenge is particularly amplified in complex architectures like ViTs. ViT encodes images as sequences of patch tokens and aggregates global context through a special classification token, the \texttt{[CLS]} token~\cite{dosovitskiy2020image}. Although this embedding provides a compact and expressive summary, it remains a black box whose internal organization is difficult to interpret~\cite{chefer2021transformer}. This interpretability gap limits our ability to build robust OOD detection mechanisms directly upon these features. 

To bridge this interpretability gap, we draw inspiration from Sparse Autoencoders (SAEs), which have recently demonstrated the ability to disentangle dense representations in large language models into sparse, semantically meaningful components~\cite{cunningham2023sparse,olshausen1997sparse}. Unlike conventional autoencoders that compress inputs into low-dimensional bottlenecks, SAEs expand the latent dimensionality while enforcing sparsity. Each input is reconstructed using only a small subset of latent neurons, revealing a structured and interpretable basis of the original representation.

We extend this paradigm to ViTs with a key modification. Traditional SAEs use ``soft'' regularization (\eg, $\ell_1$or KL penalties), which does not ensure strict sparsity, leaving many low-magnitude activations that obscure clear structural patterns. To address this, we employ a \textbf{Top-k SAE} ~\cite{gao2024scaling, makhzani2013k}, which enforces a ``hard'' sparsity constraint by activating only the $k$ most salient features for each input. This strict selection yields a stable and disjoint set of latent features for ID data, forming a clearer semantic structure. 

When trained on ViT \texttt{[CLS]} embeddings, the Top-$k$ SAE ``unfolds'' the dense representation into a structured sparse space exhibiting highly consistent, class-specific activation patterns. We formalize these patterns as \textbf{Class Activation Profiles (CAPs)}—canonical class-conditioned templates capturing the characteristic activation of each class. While ID samples preserve this structure, OOD inputs disrupt it by weakening dominant activations and diffusing sparsity patterns.

We quantify this structural discrepancy using our proposed Energy Profile Divergence (EPD), which measures how severely the sample's energy profile deviates from the characteristic energy profile defined by the class's core features. An overview of the full pipeline—from CAP construction to OOD scoring—is shown in Figure~\ref{fig1:overview}.

Our main contributions are summarized as follows:
\begin{itemize}
    \item We present the \textbf{first application of a Top-k Sparse Autoencoder} to ViT \texttt{[CLS]} tokens for OOD detection, demonstrating that hard sparsity enables interpretable and robust feature decomposition.
    \item We introduce \textbf{Class Activation Profiles (CAPs)}, an interpretable, class-conditioned structural invariant that serves as a stable reference for ID feature organization.
    \item We propose a CAP-based scoring mechanism that measures the structural divergence of core energy profiles, achieving strong robustness and competitive performance in average FPR95.
\end{itemize}


\section{Related works}
\label{sec:related}

Our study bridges three research domains: OOD detection, Vision Transformer interpretability, and sparse representation learning.

\subsection{Out-of-Distribution Detection}

OOD detection aims to identify samples that lie outside the support of the training distribution, a central challenge for reliable machine learning~\cite{yang2024generalized}. Confidence-based methods such as MSP~\cite{hendrycks2016baseline} and ODIN~\cite{liang2017enhancing} infer uncertainty from model outputs. Feature-based approaches like Mahalanobis distance~\cite{lee2018simple} and energy-based scoring~\cite{liu2020energy} evaluate distances in latent space, while others leverage gradients~\cite{huang2021importance} or outlier exposure training~\cite{hendrycks2018deep}. Generative and reconstruction-based methods use autoencoders or generative transformations to detect anomalies via reconstruction fidelity or self-supervised signals~\cite{an2015variational, golan2018deep}.

Our approach shares conceptual ground with feature-based methods but introduces a distinct perspective: we use an autoencoder not for reconstruction loss or distance measurement, but to re-parameterize the dense ViT \texttt{[CLS]} token into a sparse, interpretable latent basis. OOD detection then arises from structural disruptions in this disentangled space, rather than deviations in raw logits or entangled feature magnitudes.

\subsection{Vision Transformers and the \texttt{[CLS]} Token}

The ViT~\cite{dosovitskiy2020image} processes an image as a sequence of patch embedding tokens and aggregates contextual information through self-attention into a special \texttt{[CLS]} token~\cite{raghu2021vision}. This global embedding effectively summarizes image-level semantics but does so by entangling multiple visual attributes within a dense vector space, making it challenging to interpret or decompose~\cite{chefer2021transformer}. Recent studies have explored this embedding for OOD detection. ViM~\cite{wang2022vim} combines the ViT’s penultimate-layer feature residuals with logit information to separate ID and OOD samples, while ReAct~\cite{sun2021react} truncates activation distributions to enhance OOD separability.

However, these methods typically operate on the dense, entangled representation. Our method explicitly disentangles this representation using a SAE, recovering class-conditioned feature spaces that offer enhanced interpretability.

\subsection{Sparse Autoencoders}

SAEs are unsupervised neural models that learn overcomplete yet sparse latent representation of data, inspired by the efficient coding hypothesis in neuroscience~\cite{olshausen1997sparse}. Unlike conventional autoencoders that compress inputs into a dense bottleneck, SAEs expand the latent dimension and impose sparsity to uncover interpretable basis features. This approach has proven highly effective in large language models, where SAEs disentangle hidden activations into semantically coherent components~\cite{cunningham2023sparse}.

\subsubsection{Comparison with Other Interpretability Methods} 

Various techniques exist to interpret deep networks. Post-hoc methods like attention visualization~\cite{chefer2021transformer} or saliency maps highlight input regions relevant to a specific prediction but do not reveal the global feature vocabulary of the model. On the other hand, linear decomposition methods like Principal Component Analysis (PCA)~\cite{abdi2010principal} or Independent Component Analysis (ICA)~\cite{hyvarinen2013independent}) attempt to find global directions. However, PCA is limited to orthogonal linear transformations, and ICA assumes a linear generative process, making them ill-suilted for the highly non-linear manifolds of deep neural activation. Standard Autoencoders (AEs)~\cite{baldi2012autoencoders} generalize non-linearly but typically employ an undercomplete bottleneck, forcing compression at the cost of disentanglement.

In contrast, SAEs invert the usual objective: instead of compressing information, they expand the latent space and constrain activation sparsity. This enables the emergence of disentangled, interpretable latent factors that better capture the semantic structure of ViT representations.

\subsubsection{Soft vs Hard Sparsity}
There are two paradigms for enforcing sparsity in SAEs.

\begingroup
\setlength{\parindent}{0pt}

\textbf{Soft Sparsity.} The most common approach, popularized in deep learning~\cite{ng2011sparse}, add a soft sparsity penalty to the loss function. This penalty, typically an $\ell_1$ or a KL divergence term, encourages the average activation of hidden neurons to be low. While successful in LLM interpretability~\cite{cunningham2023sparse, elhage2022toy}, soft sparsity has limitations.

\textbf{Hard Sparsity.} A key limitation of soft sparsity is that it does not guarantee per-sample sparsity ~\cite{bricken2023monosemanticity}. An $\ell_1$-penalized model might still react to a novel OOD input with diffuse, low-magnitude activations across many features, known as the shrinkage effect, obscuring the distinction between ID and OOD. To address this, we employ the Top-k SAE~\cite{makhzani2013k, gao2024scaling}, which enforces a hard sparsity constraint. This is achieved by explicitly selecting only the $k$ features with the highest activations and zeroing out the rest.

\endgroup

In our OOD detection framework, this property is crucial: by constraining each ID sample to a stable $k$-sparse subspace, the model learns consistent, class-specific activation bases.
OOD inputs that violate these learned structural patterns naturally exhibit massive disruptions in the latent structure, providing a clear and robust signal for detection.

\section{Pattern Analysis}
\label{sec:analysis}

\begin{figure}[ht]
  \centering
   \includegraphics[width=0.98\linewidth]{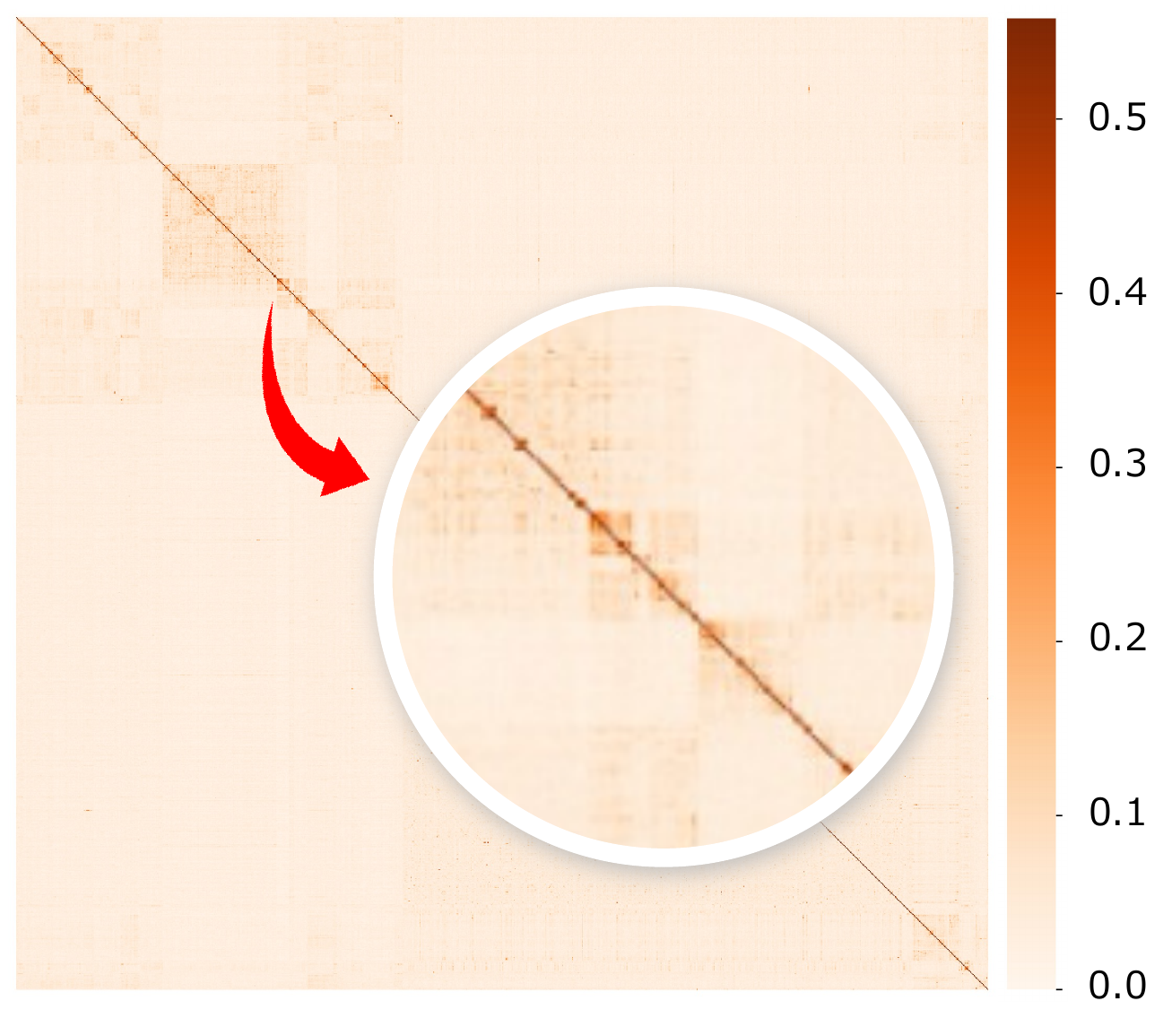}
   \caption{\textbf{Pairwise Jaccard similarity of core feature sets across all 1,000 ImageNet classes.} Each cell represents the overlap between the core feature of a class pair. The off-diagonal region indicates near-zero similarity between the core feature sets of different classes. Lighter colors indicate low similarity, while darker colors indicate high similarity. 
}
   \label{fig:jaccard}
\end{figure}

To devise an effective OOD detection method, we first analyze how our Top-$k$ SAE reshapes the dense and entangled \texttt{[CLS]} token representations. By training the SAE on \texttt{[CLS]} tokens extracted from the ImageNet-1k~\cite{deng2009imagenet} training set (ID), the dense features are decomposed into a high-dimensional, sparse latent space where semantic structures become explicit.

\subsection{Identification of Core Features in ID Classes}
\label{sec:3.1}

Our central hypothesis is that under the Top-$k$ sparsity constraint, the SAE learns a sparse basis where each class is represented by a unique set of active neurons. We posit that the dense \texttt{[CLS]} token is mapped to highly specific activation signatures that are unique for each class.

To validate this, we first define the notion of ``core features.'' For a given class $c$, we identify the subset of latent neurons with the highest mean activation on the ID training data. For our quantitative analysis of global disjointness, we instantiate this set as the top 5\% of neurons ranked by their mean activation. If these core features are truly discriminative, they should be disjoint across classes. We assess this by computing the Jaccard similarity coefficient between the core feature sets of every pair of the 1,000 ImageNet classes.

Figure~\ref{fig:jaccard} provides strong confirmation of this disjointness. The heatmap exhibits a dominant diagonal structure with near-zero off-diagonal values, indicating that the core features for different classes have minimal overlap. Occasional weak off-diagonal signals appear only between semantically related categories (\eg, different breeds of cats), reflecting shared semantic attributes. This demonstrates that the Top-$k$ SAE successfully disentangles class representations into unique, non-overlapping latent subspaces.

We further visualize this separation using UMAP~\cite{mcinnes2018umap} in Figure~\ref{fig:umap}. The 1,000 classes form tight, well-separated clusters in the projected space, confirming that the learned sparse activation patterns are not only unique but also robustly preserve the semantic identity of each class.
\subsection{OOD Activation Affinity to ID Core Features}
\label{sec:3box}

Having established that ID classes map to disjoint core features, we now investigate the behavior of OOD samples. OOD inputs, despite being unknown, are inevitably classified into one of the ID classes by the ViT backbone. Our analysis reveals that this is not a random failure but a consequence of semantic alignment in the sparse latent space: OOD samples are classified as a specific ID class because they partially activate the core features of that class.

\begin{figure}[ht]
  \centering
   \includegraphics[width=1.0\linewidth]{figs/figure4.pdf}
   \caption{\textbf{Activation affinity of OOD samples to specific ID core features.} The plots compare activations of ID samples (Blue) and misclassified OOD samples (Red) on specific core feature sets. \textbf{Top:} OOD samples from iNaturalist predicted as `Class 738 (plantpot)' show high activation on the core features of Class 738. \textbf{Bottom:} The same OOD samples show negligible activation on the core features of an unrelated class (`Class 989, rosehip'). This illustrates that OOD samples are not random but structurally align with their predicted class.
}
   \label{fig:activation}
\end{figure}

\begin{figure*}[t]
  \centering
   \includegraphics[width=0.95\linewidth]{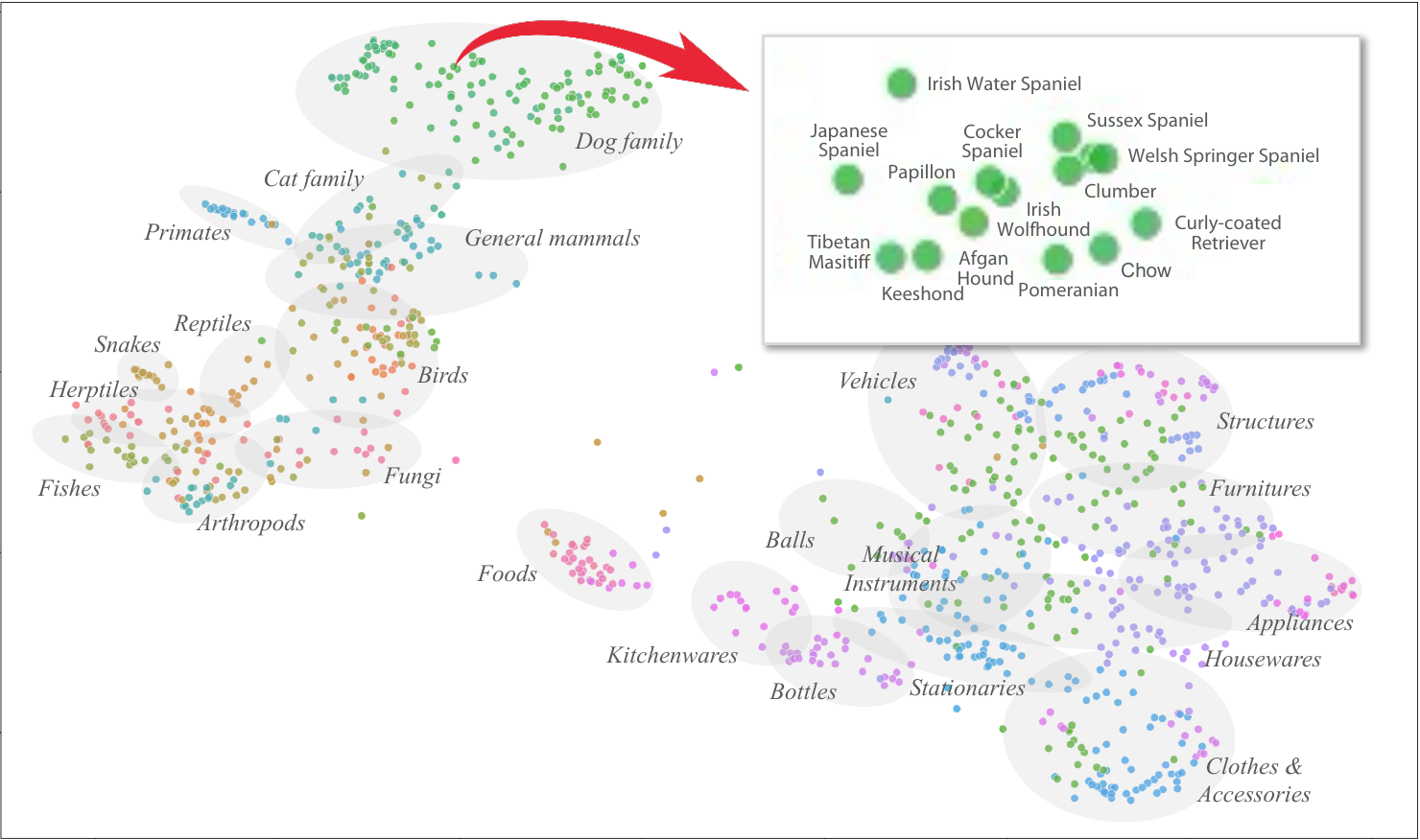}
   \caption{\textbf{UMAP visualization clustering the mean sparse activation vectors of all 1,000 ID classes.} Each point represents the centroid of a class in the SAE latent space. The formation of tight, well-separated clusters confirms that the sparse features robustly capture the distinct semantic identity of each class.
}
   \label{fig:umap}
\end{figure*}

We visualize this phenomenon in Figure~\ref{fig:activation}, using the iNaturalist dataset~\cite{van2018inaturalist} as a representative OOD dataset. The figure compares the activation magnitudes on the core feature sets. To ensure visual clarity, we plot the top-20 dominant core features. The top panel shows OOD samples misclassified as `Class 738 (plantpot)'. Noticeably, these samples exhibit significant, non-zero activations on the core features of Class 738, mirroring the behavior of true ID samples. Conversely, the bottom panel shows the same OOD samples evaluated on the core features of an unrelated class (`Class 989, rosehip'), where activations collapse to near-zero.

This explains the mechanism of misclassification: the OOD sample is routed to a specific ID class because its underlying features align better with that class's core features than any other. This behavior is consistently observed across other datasets, including Textures~\cite{cimpoi2014describing} and OpenImage-O~\cite{kuznetsova2020open}.

To quantify this effect globally, Figure~\ref{fig:3box} aggregates these statistics across all classes. We compare: (1) ID samples on their \textit{true} core features (Blue, Left), (2) OOD samples on their \textit{matched} class's core features (Red, Middle), and (3) OOD samples on all \textit{other} class features (Red, Right).

The clear separation between the (2) and (3) provides statistical evidence for our hypothesis. OOD samples do not activate neurons randomly; they systematically align with the core features of the single ID class they resemble most. However, a crucial distinction remains: as seen in the gap between the Blue and Middle-Red boxes, the magnitude of this activation is weaker than that of true ID samples. We explore this vital structural difference in the next section.
\subsection{Difference of Activation Profiles in OOD}
\label{sec:profile}

\begin{figure}[ht]
  \centering
   \includegraphics[width=1.0\linewidth]{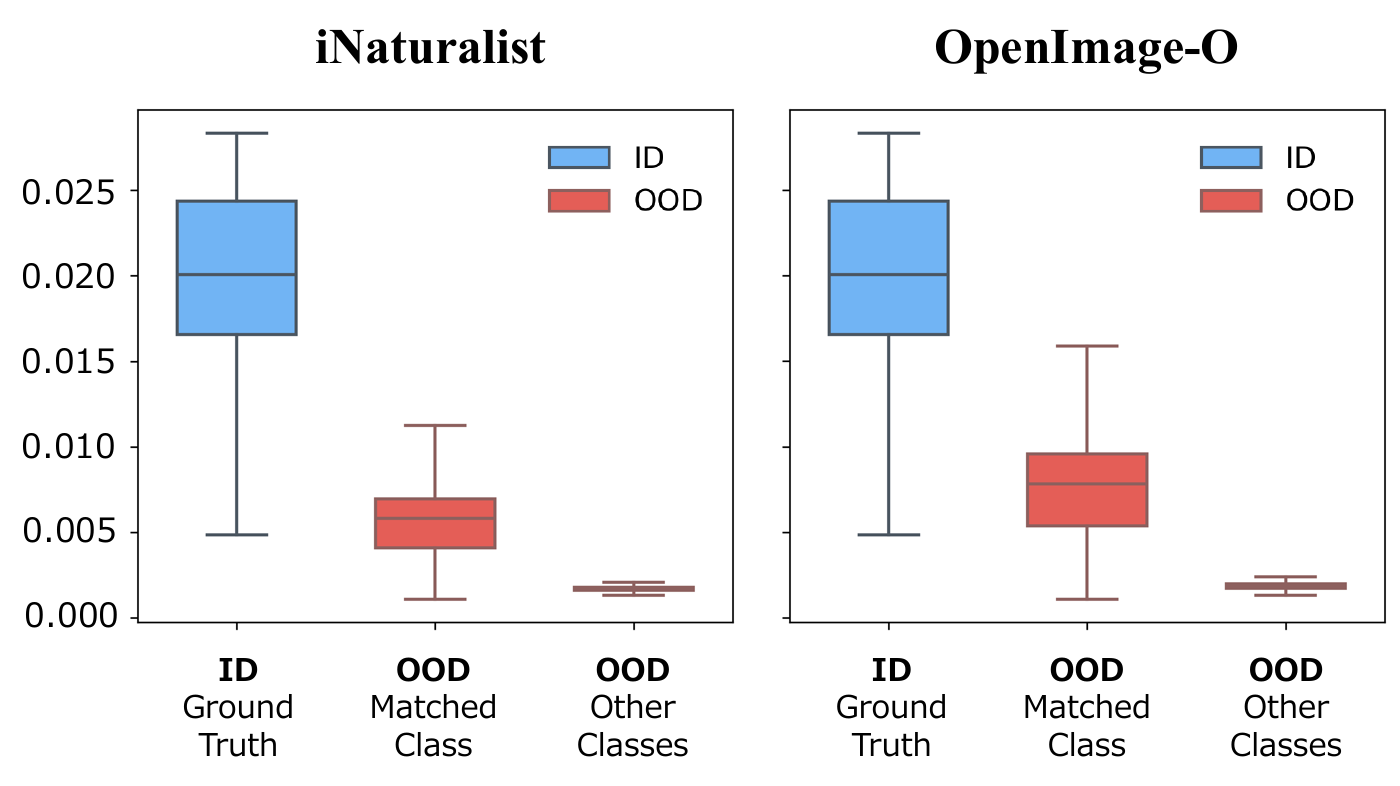}
   \caption{\textbf{Global statistics of activation intensity.} We aggregate mean activations on core indices across all classes for two OOD datasets: iNaturalist (Left) and OpenImage-O (Right). \textbf{ID Ground Truth (Blue, Left):} Strong activation on ground-truth core features. \textbf{OOD Matched (Red, Middle):} OOD samples activate the core features of their \textit{predicted} class, but with lower intensity than ID. \textbf{OOD Other Classes (Red, Right):} OOD samples show minimal activation on unrelated classes.
}
   \label{fig:3box}
\end{figure}

Our analysis in the Section~\ref{sec:3box} demonstrates that OOD samples mimic the core features of ID classes. However, Figure~\ref{fig:3box} suggests that while they hit the correct features, they fail to match the intensity of ID samples.

\begin{figure}[h]
  \centering
   \includegraphics[width=1.0\linewidth]{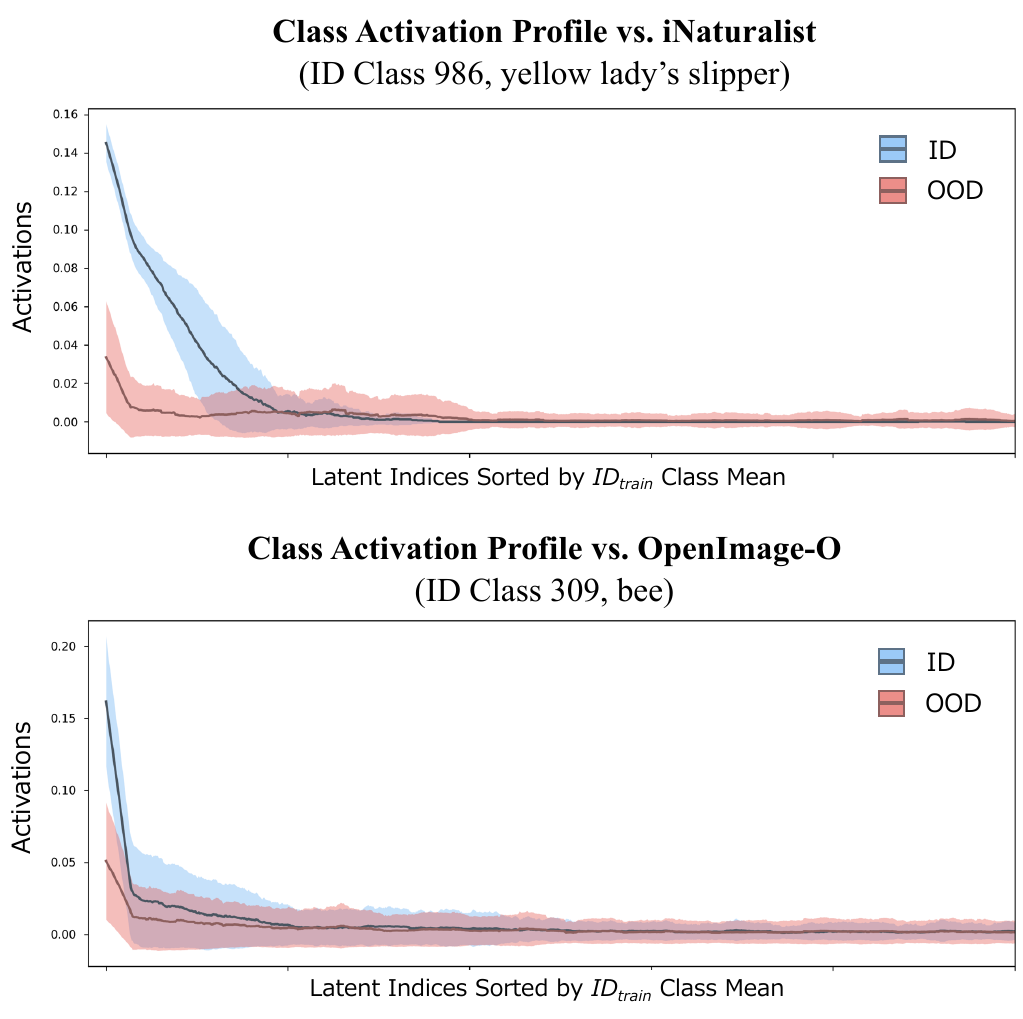}
   \caption{\textbf{Structural differences in ID and OOD activation.} We compare the mean activation profiles of ID samples (Blue) and misclassified OOD samples (Red) sorted by the ID CAP. \textbf{Top:} iNaturalist vs. ID (Class 986). \textbf{Bottom:} OpenImage-O vs. ID (Class 309). In both cases, ID samples maintain a sharp, concentrated head, whereas OOD samples exhibit a flattened, diffused profile. Shaded regions indicate variance.
}
   \label{fig:IDvsOOD}
\end{figure}
We formalize this discrepancy by analyzing the \textbf{Class Activation Profile (CAP)}: the full $D_{latent}$-dimensional mean activation vector for a class. This serves as a canonical template of the ``ideal'' activation structure for that class.

Figure~\ref{fig:IDvsOOD} visualizes the activation profiles of ID samples versus OOD samples predicted as the same class against this canonical template.
This comparison reveals a fundamental structural disruption.
ID samples (Blue) exhibit a sharp, high-energy head where activation is highly concentrated, followed by a rapid drop-off. In contrast, OOD samples (Red) show a significantly flatter profile. They fail to activate the critical features in the CAP's sorted head with the same intensity.

This structural deviation is the cornerstone of our OOD detection method. Even when an OOD sample fools the classifier by activating the correct subset of core features, it fails to replicate the precise activation pattern—the shape of the distribution—of the ID data. While Figure~\ref{fig:IDvsOOD} illustrates this for iNaturalist~\cite{van2018inaturalist} and OpenImage-O~\cite{kuznetsova2020open}, this pattern is consistent across diverse OOD benchmark datasets. This insight directly motivates our proposed scoring function, Energy Profile Divergence (EPD), which is explicitly designed to quantify this structural disruption in the energy allocation profile.
\section{Experiments}
\label{sec:exp}
\subsection{Experimental Setup}
\label{sec:exp_setup}
\begingroup
\setlength{\parindent}{0pt}
\textbf{Benchmark.} In our experiments, we adopt \textbf{OpenOOD v1.5}~\cite{zhang2023openood}, the latest version of the comprehensive OOD detection benchmark framework, to ensure a fair evaluation against existing works. The ID data for all experiments is ImageNet-1k~\cite{deng2009imagenet}. The OOD evaluation was conducted using a suite of five datasets, organized into two groups based on their semantic distance to ImageNet-1k: near-OOD (SSB-hard~\cite{vaze2021open}, NINCO~\cite{bitterwolf2023ninco}) and far-OOD (iNaturalist~\cite{van2018inaturalist}, Textures~\cite{cimpoi2014describing}, OpenImage-O~\cite{kuznetsova2020open}).

\textbf{Backbone Architectures.} 
Our primary experiments are conducted using a pre-trained ViT-B/16 as the backbone. To verify the generalizability, we also evaluate it on the Swin Transformer~\cite{liu2021swin} and DINOv2~\cite{oquab2023dinov2}.

\textbf{OOD Scoring.} As motivated in Section~\ref{sec:profile}, detecting OOD requires quantifying \textit{structural} deviation of activation profiles. To measure this, we first define the CAP for a class $c$, denoted as $\bar{\mathbf{h}}^c \in \mathbb{R}^{D_{latent}}$, which is the mean activation vector of all ID training samples belonging to that class. Let $\mathbf{h}^s \in \mathbb{R}^{D_{latent}}$ be the sparse activation vector of a given test sample $s$. 

We define $M^c$ as the array of indices corresponding to the $L$ largest mean activation values in the CAP $\bar{\mathbf{h}}^c$. This parameter $L$ defines the number of core features we analyze for each class. We then filter both the CAP and the sample's activation vector using this index set $M^c$ to create two $L$-dimensional core activation vectors, $\mathbf{C}$ and $\mathbf{S}$
\begin{equation}
\begin{aligned}
\mathbf{C}_i = \bar{\mathbf{h}}^c_{M_i^c},
\quad 
\mathbf{S}_i = \mathbf{h}^s_{M_i^c},
\quad \forall i \in {1,\dots,L}
\end{aligned}
\end{equation}

We interpret these $L$-dimensional vectors as energy distribution profiles. The total $L_1$ norm represents the total energy allocated within this core subspace, while $L_1$ normalization isolates the \textit{proportional allocation} of this energy.
This normalization effectively projects the core activation vector onto a $(L-1)$-dimensional simplex. This simplex is the class-specific geometric space we analyze, isolating the semantic ``shape'' of the class's core feature hierarchy from its overall scale.

We define their normalized energy distribution profiles, $\mathbf{P}$ (from the sample $\mathbf{S}$) and $\mathbf{Q}$ (from the class CAP $\mathbf{C}$), as:
\begin{equation}
\mathbf{P}_i = \frac{\mathbf{S}_{i}}{\sum\limits_{i=1}^{L}{\mathbf{S}_i}}, \quad 
\mathbf{Q}_i = \frac{{\mathbf{C}}_{i}}{\sum\limits_{i=1}^{L}{\mathbf{C}_i}},  \quad \forall i \in {1,\dots,L}
\end{equation}
Within this core feature space, the reference profile $\mathbf{Q}$ serves as the geometric anchor point or centroid on the simplex, representing the stable structure of the ID class. The test sample's profile $\mathbf{P}$ is a new point projected into this same space. We quantify the discrepancy between them using our metric, the \textbf{Energy Profile Divergence (EPD)}:
\begin{equation}
    \text{EPD Score} = D_{\text{KL}}(\mathbf{P} \parallel \mathbf{Q}) = \sum_{i=1}^L \mathbf{P_i} \log \left(\frac{\mathbf{P_i}
    }{\mathbf{G_i}}\right)
    \label{eq:epd_score}
\end{equation}
This score, formally equivalent to the KL Divergence $D_{\text{KL}}(\mathbf{P} \parallel \mathbf{Q})$~\cite{kullback1951information}, provides a principled measure of geometric divergence within this class-specific subspace. A high EPD score signifies that the test sample $\mathbf{P}$ has ``drifted'' far from the characteristic anchor point $\mathbf{Q}$, indicating that its energy allocation fundamentally disrupts the ID class's core feature hierarchy.

\begingroup
\setlength{\parindent}{0pt}
\textbf{Metrics.} 
We report using two primary metrics: the Area Under the Receiver Operating Characteristic (AUROC) curve (higher is better) and the False Positive Rate at 95\% True Positive Rate (FPR95) (lower is better). While AUROC provides a general measure of separability across all thresholds, FPR95 is the critical metric for reliable deployment in safety-sensitive applications, as it bounds the worst-case failure rate on anomalies~\cite{hendrycks2021unsolved}. It directly measures the OOD false positive rate under the constraint that 95\% of ID samples are correctly accepted.
\endgroup

\subsection{SAE Architecture and Training}
\label{sec:sae_arch}

We adopt a Top-$k$ SAE~\cite{makhzani2013k}, an overcomplete, single-hidden-layer autoencoder. Unlike traditional SAEs that use $\ell_1$ or KL penalties to encourage average sparsity~\cite{ng2011sparse}, our model enforces per-sample hard sparsity.

The model is trained to optimize a composite loss function designed to balance accurate reconstruction with maintaining robust feature activation. The total loss is:
\begin{equation}
\mathcal{L}_{\text{Total}} = \mathcal{L}_{\text{Recon}} + \alpha \mathcal{L}_{\text{AuxK}}
\label{eq:sae_loss}
\end{equation}
Reconstruction Loss ($\mathcal{L}_{\text{Recon}}$) is the Mean Squared Error (MSE) between the normalized input \texttt{[CLS]} token and its reconstruction. To mitigate ``neuron dead''—a common issue in Top-$k$ models where some neurons never win the $k$-competition—we employ an auxiliary loss, $\mathcal{L}_{\text{AuxK}}$~\cite{gao2024scaling}. This loss encourages dead neurons to participate in reconstructing the residual error, weighted by $\alpha$.

This hard-sparsity architecture is fundamental to our hypothesis. Soft $\ell_1$ sparsity can still permit dense, low-amplitude activations in response to novel OOD inputs, corrupting the signal. In contrast, the Top-k SAE acts as a structural bottleneck, forcing a sparse representation. This amplifies the dissimilarity between the stable ID patterns (CAPs) and the disrupted OOD patterns, making the structural deviation quantified by our EPD a highly robust OOD signal.

\subsection{Hyperparameter Rationale}
\label{sec:hyperparams}
Our hyperparameters were selected to balance representational capacity with semantic specificity:

\begingroup
\setlength{\parindent}{0pt}

\textbf{Latent Dimension} ($D_{latent}=7680$)\textbf{.} Our latent dimension is 10x the ViT-B/16 \texttt{[CLS]} token dimension ($D_{in}=768$). This overcomplete ratio is for providing sufficient capacity for the features to disentangle into a sparse basis~\cite{cunningham2023sparse}.

\textbf{Sparsity Level} ($k=128$)\textbf{.}  We empirically identified $k=128$ as the optimal operating point. This value is significantly smaller than $D_{in}$ (768), enforcing a strong representational bottleneck that compels the model to select only the most salient features for reconstruction, achieving a 6x dimensionality reduction per-sample.

\textbf{Activation Head Size} ($p=15\%$)\textbf{.} From a 99\%-firing analysis, we observed that each class activates approximately 18.8\% of the latent dimensions on average (std.\ 4.6\%), implying a meaningful activation band of roughly 14–23\%. Guided by 
this empirical range and validated through performance sweeps, we set the activation head size to the corresponding optimum.

These parameters were finalized to achieve a balanced trade-off between sensitivity to OOD detection performance and robustness across diverse benchmarks. See the Appendix for a detailed results.

\endgroup

\subsection{Computational Cost}
We benchmarked the Top-$k$ SAE training cost on a single NVIDIA RTX 4080 GPU: 100 epochs over ImageNet-1k (~1.28M samples, batch size 4096) completed in approximately 17 minutes (10.18 sec/epoch). This constitutes a one-time setup cost, as the SAE is subsequently frozen during inference, requiring no retraining or fine-tuning, ensuring the computational efficiency of our method.

\subsection{Results on ViT}
\label{sec:vit_results}

Table \ref{tab:result-vit-summary} compares our method against prominent approaches on a ViT-B/16 backbone. Our method achieves the best average FPR95 (40.96\%) among all compared methods, demonstrating strong robustness in OOD detection. As discussed in Section~\ref{sec:exp_setup}, minimizing FPR95 is paramount for safety. Concurrently, our method secures a highly competitive separability, securing the second-best overall AUROC (87.26\%). Table \ref{tab:result-vit-ours} presents the per-dataset AUROC and FPR95 scores for our method, where our method remains highly competitive, particularly achieving superior FPR95 on SSB-Hard and Textures. Detailed results for the other benchmark methods are provided in the Appendix.

\begin{table}[ht]
  \centering
    \resizebox{\columnwidth}{!}{
        \begin{tabular}{lcccccc} 
            \toprule
            \multirow{2}{*}{\textbf{Method}} & 
            \multicolumn{3}{c}{\textbf{FRP95}} & 
            \multicolumn{3}{c}{\textbf{AUROC}} \\
            \cmidrule(lr){2-4} \cmidrule(lr){5-7}
            & \textbf{Near-OOD} & \textbf{Far-OOD} & \textbf{Average} & \textbf{Near-OOD} & \textbf{Far-OOD} & \textbf{Average} \\ 
            \midrule
            ASH       & 94.45  & 96.77 & 95.84 & 53.20 & 51.56 & 52.22 \\
            GEN       & 70.78 & 32.23 & 47.65 & 76.30 & 91.35 & 85.33 \\
            GradNorm  & 94.72 & 92.64 & 93.47 & 53.68 & 41.74 & 46.52 \\
            KNN       & 70.47 & 31.93 & 47.35 & 74.11 & 90.81 & 84.13 \\
            MDS       & \underline{66.11} & 29.97 & \underline{44.43} & \underline{79.04} & \underline{92.60}  & \underline{87.18} \\
            MSP       & 81.88 & 51.66 & 63.75 & 73.53 & 86.04 & 81.03 \\
            OpenMax   & 88.78 & 55.49 & 68.81 & 73.64 & 89.27 & 83.02 \\
            ReAct     & 84.48 & 53.92 & 66.15 & 69.27 & 85.69 & 79.12 \\
            RMDS      & \underline{65.38} & \underline{28.75} & \underline{43.40} & \textbf{80.09} & \underline{92.60} & \textbf{87.60} \\
            SHE       & 70.88 & \underline{27.14} & 44.63 & 76.11 & 92.42  & 85.90 \\
            ViM       & 73.75 & 29.20 & 47.02 & 77.03 & \textbf{92.84} & 86.51 \\
            TempScale & 84.63 & 53.77 & 66.11 & 73.17 & 86.32 & 81.06 \\
            \rowcolor{black!10}
            \textbf{Ours} & \textbf{65.23} & \textbf{24.77} & \textbf{40.96} & \underline{78.98} & \underline{92.78} & \underline{87.26} \\
            \bottomrule
        \end{tabular}
    }
  \caption{OOD detection performance on ViT-B/16 (ID Acc: 81.14\%). Results are averaged across Near-OOD and Far-OOD datasets. Best results are in bold; second and third-best are underlined. Our method achieves the best average FPR95.}
  \label{tab:result-vit-summary}
\end{table}

\begin{table}[ht]
  \centering
  \resizebox{\columnwidth}{!}{
    \begin{tabular}{l c c c c c}
    \toprule
    \textbf{Metric} & \textbf{SSB hard} & \textbf{NINCO} & \textbf{iNaturalist} & \textbf{Textures} & \textbf{OpenImage-O} \\
    \midrule
    \textbf{FPR95}    & \underline{82.41} & \underline{48.06} & \underline{17.84} & \underline{30.44} & \textbf{26.03} \\
    \textbf{AUROC}    & \underline{72.21} & \underline{85.74} & 95.17 & \underline{91.06} & \underline{92.12} \\
    \bottomrule
    \end{tabular}
  }
  \caption{Detailed our OOD detection performance on ViT-B/16. Best results are in bold; second and third-best are underlined.}
  \label{tab:result-vit-ours}
\end{table}

Our results demonstrate strong robustness across diverse OOD scenarios. Our approach also achieves leading FPR95 on Far-OOD datasets and maintains competitive performance on notoriously difficult Near-OOD benchmarks.

Robustness on individual datasets is further detailed in the Appendix. Particularly, our method records the best FPR95 performance on OpenImage-O (26.03\%) and maintains top-tier performance on the challenging datasets like NINCO (48.06\%), outperforming strong baselines.

These results support our core hypothesis: an OOD signal derived from structural deviation within a learned sparse feature space provides a reliable detection mechanism. By enforcing hard sparsity, the Top-$k$ SAE learns a discriminative latent space where our EPD effectively detects both Far-OOD samples (failing to activate core features) and Near-OOD samples (activating a conflicting pattern).

\subsection{Results on Other Backbones}
\label{sec:other_backbones}
Our method also achieves competitive results on the Swin-T backbone, showing strong robustness in FPR95 compared to existing methods (Table \ref{tab:result-swint-summary}). However, its overall performance is lower than that on ViT-B/16. We attribute this gap not to a fundamental limitation of Swin-T, but to how its architectural design interacts with our approach. Our method assesses OOD likelihood by comparing activation-profile shapes between a sample and its predicted class’ CAP. Swin-T’s window-based local attention and hierarchical aggregation yield feature embeddings with weaker global consistency and less sharply defined activation heads, which directly reduces the separability achieved through profile-shape alignment.

\begin{table}[ht]
  \centering
    \centering
    \resizebox{\columnwidth}{!}{
        \begin{tabular}{lcccccc} 
            \toprule
            \multirow{2}{*}{\textbf{Method}} & 
            \multicolumn{3}{c}{\textbf{FRP95}} & 
            \multicolumn{3}{c}{\textbf{AUROC}} \\
            \cmidrule(lr){2-4} \cmidrule(lr){5-7}
            & \textbf{Near-OOD} & \textbf{Far-OOD} & \textbf{Average} & \textbf{Near-OOD} & \textbf{Far-OOD} & \textbf{Average} \\ 
            \midrule
            ASH        & 94.60 & 94.78 & 94.71 & 46.64 & 44.34 & 45.26 \\
            GEN        & \textbf{63.72} & 32.76 & 45.15 & \textbf{78.97} & 91.00 & \underline{86.19} \\
            GradNorm   & 93.72 & 96.60 & 95.45 & 47.72 & 35.68 & 40.49 \\
            KNN        & 71.61 & 34.41 & 46.28 & 71.69 & 89.36 & 82.45 \\
            MDS        & 68.64 & \underline{30.75} & \underline{42.74} & 75.24 & \underline{91.46} & \underline{86.92} \\
            MSP        & 70.69 & 49.32 & 52.27 & 76.72 & 89.65 & 84.47 \\
            OpenMax    & 79.20 & 52.28 & 63.06 & 76.61 & 88.54 & 83.96 \\
            ReAct      & 72.55 & 42.62 & 56.30 & 75.74 & 88.99 & 83.69 \\
            RMDS       & \underline{66.28} & \underline{29.33} & \textbf{41.37} & \underline{78.36} & \textbf{92.98} & \textbf{88.36} \\
            SHE        & 76.28 & 45.82 & 58.00 & 76.75 & 89.89 & 84.63 \\
            TempScale  & 72.57 & 50.01 & 59.03 & \underline{76.96} & 86.75 & 82.83 \\
            \rowcolor{black!10}
            \textbf{Ours} & \underline{68.51} & \textbf{26.97} & \underline{43.99} & 76.76 & \underline{92.21} & 86.03 \\
            \bottomrule
        \end{tabular}
    }
  \caption{OOD detection performance on Swin Transformer (ID Acc: 81.60\%). Results are averaged across Near-OOD and Far-OOD datasets. Best results are in bold; second and third-best are underlined.}
  \label{tab:result-swint-summary}
\end{table}

In contrast, our method demonstrates strong performance when applied to DINOv2, which is explicitly optimized for object-centric global representations. The resulting CAPs are significantly sharper and more stable than those of Swin-T, leading to a substantial reduction in FPR95. This confirms that our structural detection mechanism effectively leverages the global coherence of the latent space, performing best when the backbone naturally minimizes intra-class feature variance. Detailed results are included in the Appendix.

\subsection{Ablation: Scoring Metric}
\label{sec:metric_ablation}
To validate our choice of EPD, we compare it against Euclidean and cosine distance applied within the same sparse CAP framework on ViT-B/16. EPD outperforms both alternatives across all evaluation splits; detailed results are provided in the Appendix.
\section{Conclusion}
\label{sec:conclusion}

We demonstrated that the ViT \texttt{[CLS]} token can be interpreted by decomposing it into a structured sparse basis. To the best of our knowledge, this work represents the first application of a Top-$k$ SAE to unlock these features specifically for OOD detection.

A core contribution is the identification of CAPs as stable structural invariants for ID classes. Our analysis revealed that OOD samples fail to replicate the precise shape of these profiles. 

By quantifying this structural mismatch via EPD, our method achieved strong performance in average FPR95. This confirms that analyzing the distributional shape of activations provides a robustness advantage over conventional magnitude-based approaches.

Beyond performance, our framework is computationally efficient—requiring only a lightweight, one-time SAE train—and interpretable, establishing structural alignment as a robust foundation for reliable vision systems

\noindent\textbf{Limitations and Future Work.} While we established the interpretability of sparse patterns at the class level via CAPs, we did not conduct a granular feature-level visualization to identify the specific semantic concepts captured by individual latent neurons. We leave this fine-grained analysis for future work, as well as extending our structural framework to other modalities and architectures.

\noindent\textbf{Acknowledgments.}
This work was supported by the National Research Foundation of Korea (NRF) grant (No. RS-2024-00453301) and the Institute for Information \& communications Technology Planning \& Evaluation (IITP) grant (No. RS-2025-25442469) funded by the Korea government (MSIT).

{
    \small
    \bibliographystyle{ieeenat_fullname}
    \bibliography{main}
}

\clearpage
\setcounter{page}{1}
\setcounter{section}{0}

\onecolumn

\begin{center}
  
  {\Large \bfseries Sparsity as a Key: Unlocking New Insights from Latent Structures\\[0.3em]for Out-of-Distribution Detection}
  
  \vspace{0.7em}
  
  {\Large Supplementary Material \par}
  
  \vspace{0.7em}
\end{center}

\setcounter{section}{0}

\appendix

\section{Activation Affinity of OOD Samples to Specific ID Core Features}
\begin{figure}[h]
    \centering
    \begin{subfigure}[b]{1.0\linewidth}
        \centering
        \includegraphics[width=0.90\linewidth]{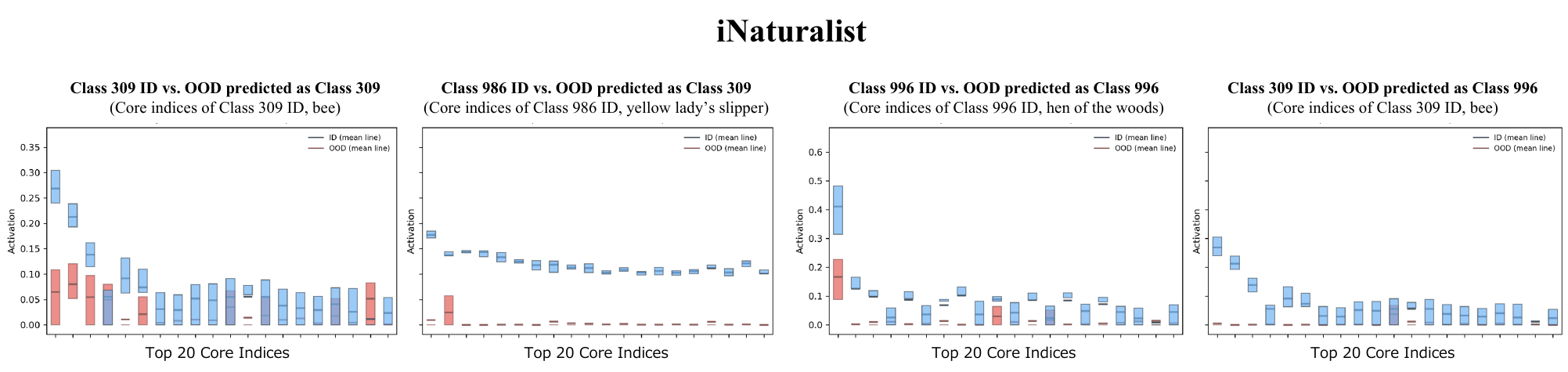}
        \label{fig:activation_appendix7-1}
    \end{subfigure}
    \vspace{0.5cm}
    \begin{subfigure}[b]{1.0\linewidth}
        \centering
        \includegraphics[width=0.90\linewidth]{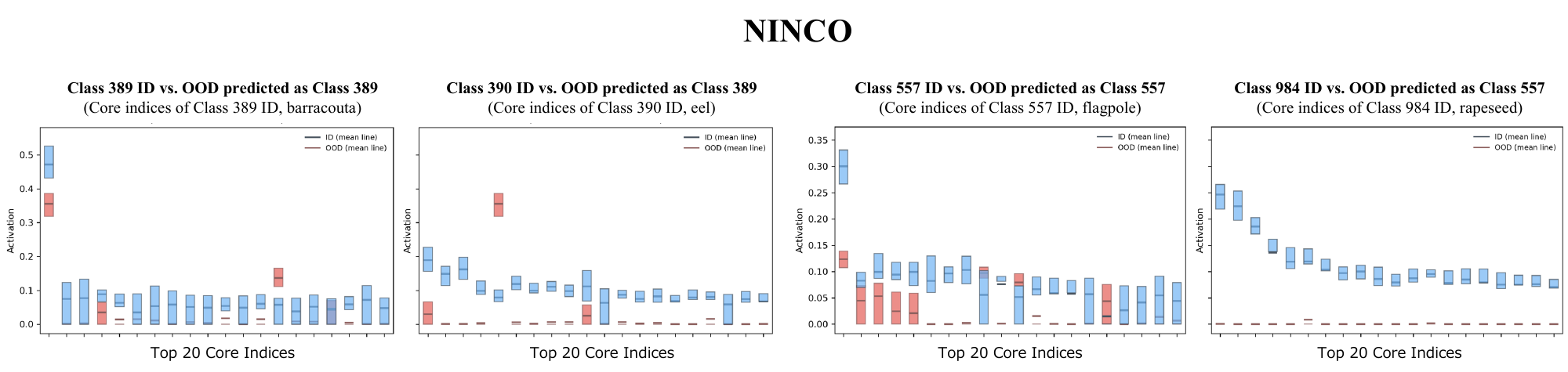}
        \label{fig:activation_appendix7-2}
    \end{subfigure}
    \vspace{0.5cm}
    \begin{subfigure}[b]{1.0\linewidth}
        \centering
        \includegraphics[width=0.90\linewidth]{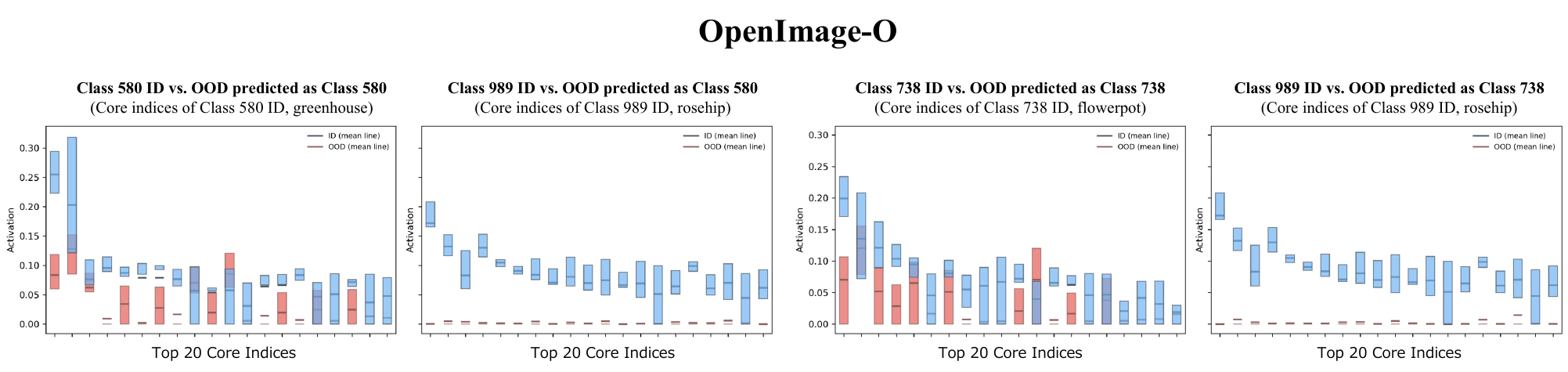}
        \label{fig:activation_appendix7-3}
    \end{subfigure}
    \vspace{0.5cm}
    \begin{subfigure}[b]{1.0\linewidth}
        \centering
        \includegraphics[width=0.90\linewidth]{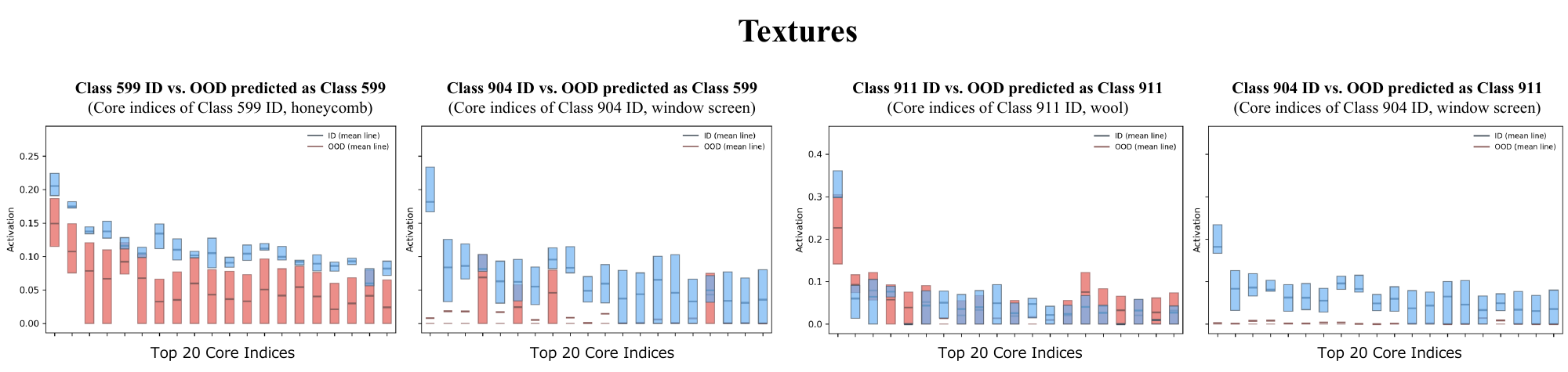}
        \label{fig:activation_appendix7-4}
    \end{subfigure}
    
    \caption{\textbf{Core feature activation by different OOD datasets.} 
    This figure presents two pairs of graphs for each dataset. Each pair shows an OOD sample (Red) compared against ID samples (Blue) using two class feature sets. 
    \textbf{Left panel (in each):} The OOD sample consistently exhibits high activation on the core features of its predicted ID class.
    \textbf{Right panel (in each):} The same OOD sample demonstrates negligible activation on the core features of an unrelated ID class. This pattern confirms the robust semantic alignment driving classifier predictions, irrespective of the OOD source dataset.}
    \label{fig:activation_appendix}
\end{figure}

To demonstrate that the semantic alignment of OOD samples is a general phenomenon—a key finding discussed in Section \ref{sec:3box}—Figure \ref{fig:activation_appendix} presents additional evidence drawn from diverse OOD datasets, including iNaturalist, NINCO, OpenImage-O, and Textures. As evident in the figure, the pattern remains remarkably consistent. In every illustrated case, the OOD sample (Red) exhibits significant, non-zero activation levels when evaluated against the core features of its misclassified ID class (Left panels in each). Conversely, the same OOD sample consistently demonstrates negligible activation when processed against the core features belonging to an entirely unrelated ID class (Right panels in each). This observation strongly reinforces our hypothesis that the classifier's decisions are driven by structural and semantic alignments in the sparse latent space, rather than resulting from random classification failures.

\vspace{2.0em}

\section{Global Statistics of Activation Intensity}
\begin{figure}[h]
  \centering
   \includegraphics[width=0.85\linewidth]{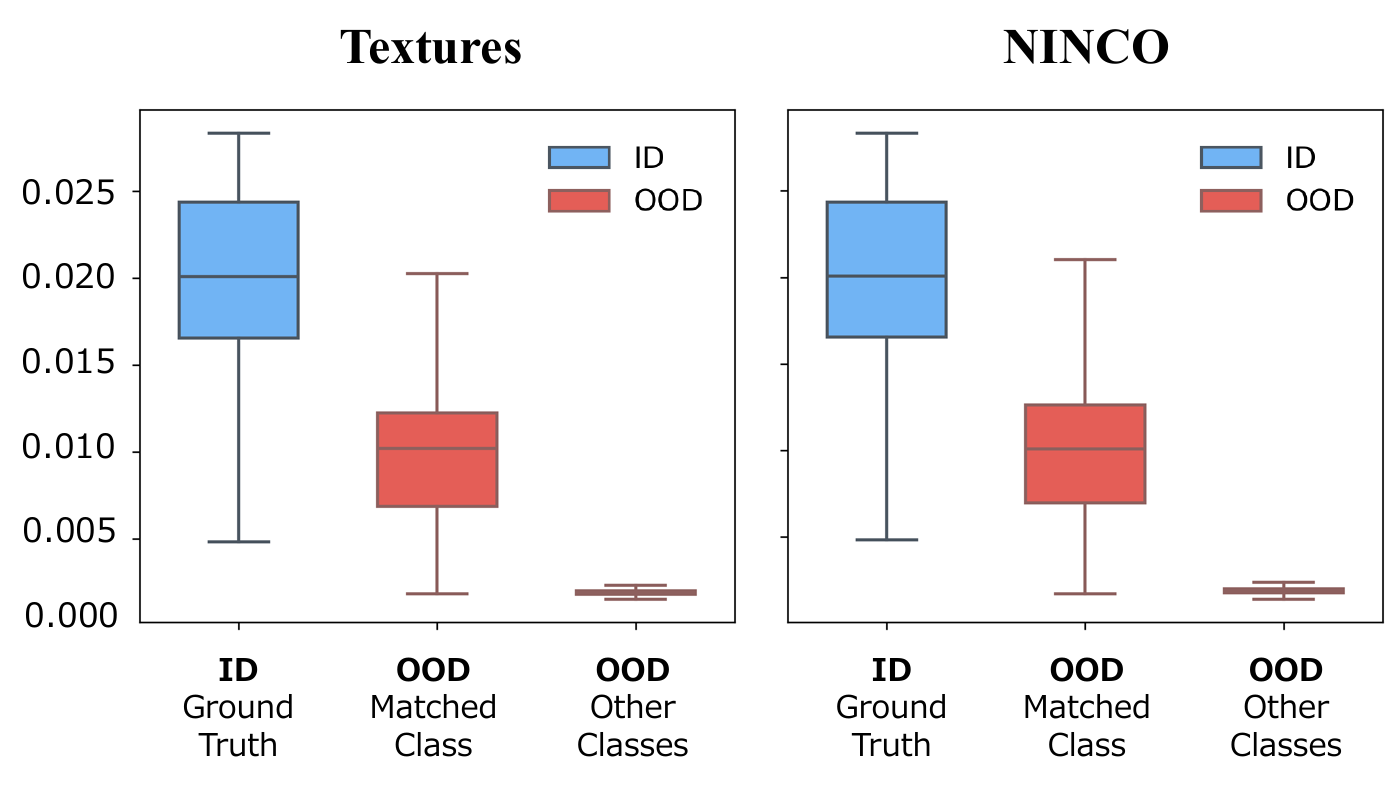}
   \caption{\textbf{Global statistics of core feature activation intensity.} We aggregate mean activation values on core indices across all classes for OOD datasets. \textbf{ID Ground Truth (Blue, Left):} Demonstrates strong activation on the core features corresponding to the ground-truth class. \textbf{OOD Matched Class (Red, Middle):} OOD samples activate the core features of their \textit{predicted} ID class, though with consistently lower intensity than true ID samples. \textbf{OOD Other Classes (Red, Right):} OOD samples exhibit minimal activation on the core features of unrelated ID classes.}
   \label{fig:3box_appendix}
\end{figure}

As a supplement to the analysis Figure \ref{fig:3box} in Section \ref{sec:3box}, Figure \ref{fig:3box_appendix} demonstrates that our findings are not limited to specific examples but represent a general phenomenon across diverse OOD datasets. The aggregated statistics consistently show that OOD samples systematically activate their \textit{matched} class features (Red, Middle) but not \textit{other} class features (Red, Right). This clear separation provides another strong evidence supporting our hypothesis that this specific activation is the reason why the sample is classified into that particular ID class. Furthermore, this OOD activation magnitude (Red, Middle) is consistently weaker than that of true ID samples (Blue, Left).

\clearpage

\section{Structural Differences in ID and OOD Activation}

\begin{figure}[h]
    \centering
    \begin{subfigure}[b]{1.0\linewidth}
        \centering
        \includegraphics[width=0.90\linewidth]{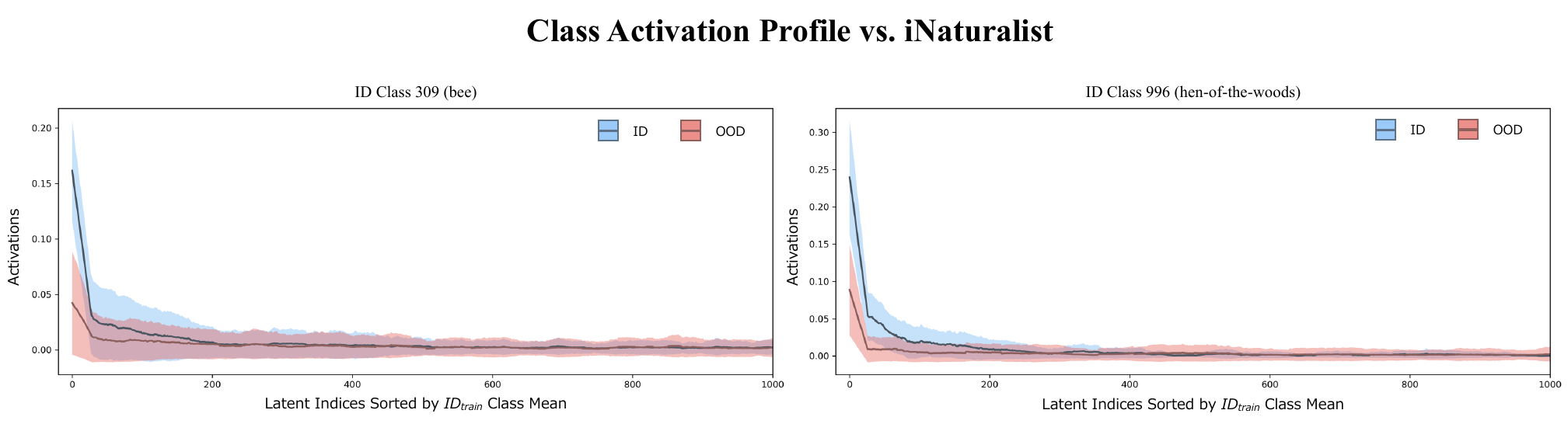}
        \label{fig:activation_appendix9-1}
    \end{subfigure}
    \vspace{0.5cm}
    \begin{subfigure}[b]{1.0\linewidth}
        \centering
        \includegraphics[width=0.90\linewidth]{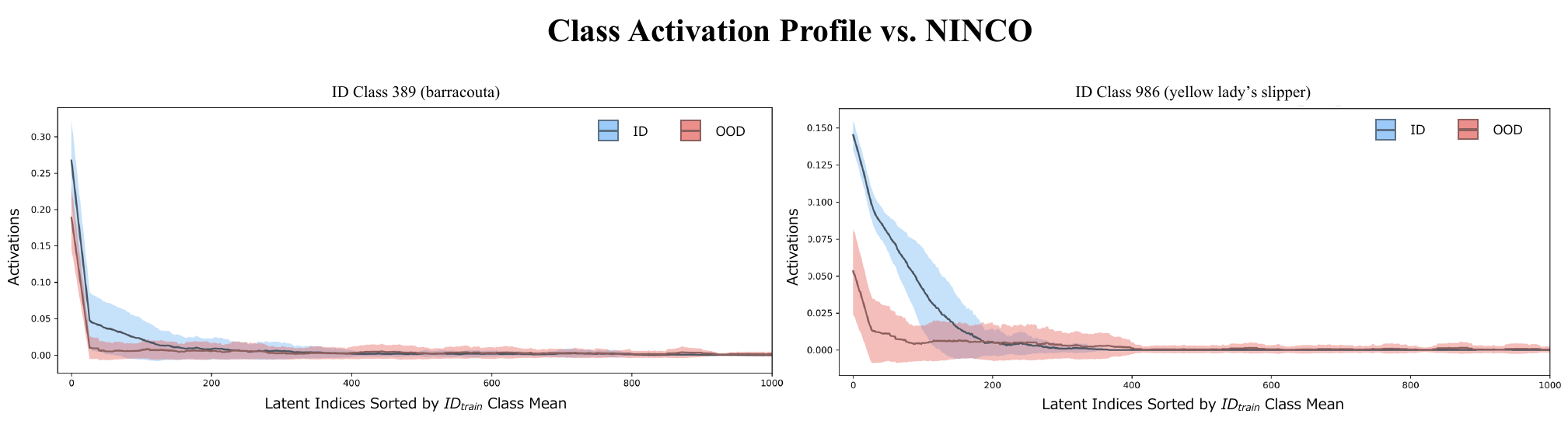}
        \label{fig:activation_appendix9-2}
    \end{subfigure}
    \vspace{0.5cm}
    \begin{subfigure}[b]{1.0\linewidth}
        \centering
        \includegraphics[width=0.90\linewidth]{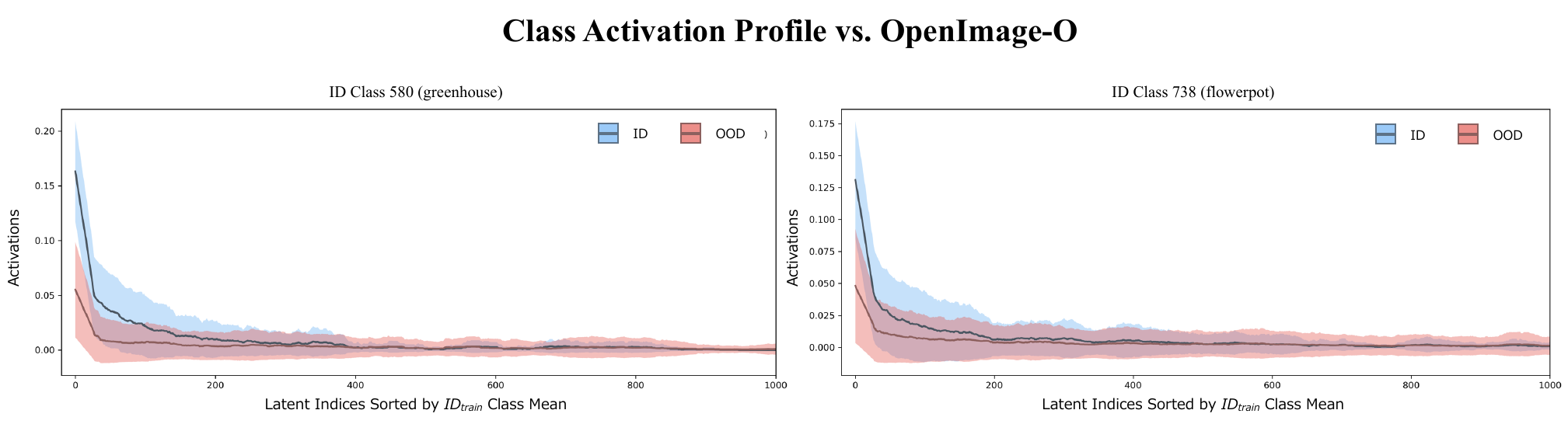}
        \label{fig:activation_appendix9-3}
    \end{subfigure}
    \vspace{0.5cm}
    \begin{subfigure}[b]{1.0\linewidth}
        \centering
        \includegraphics[width=0.90\linewidth]{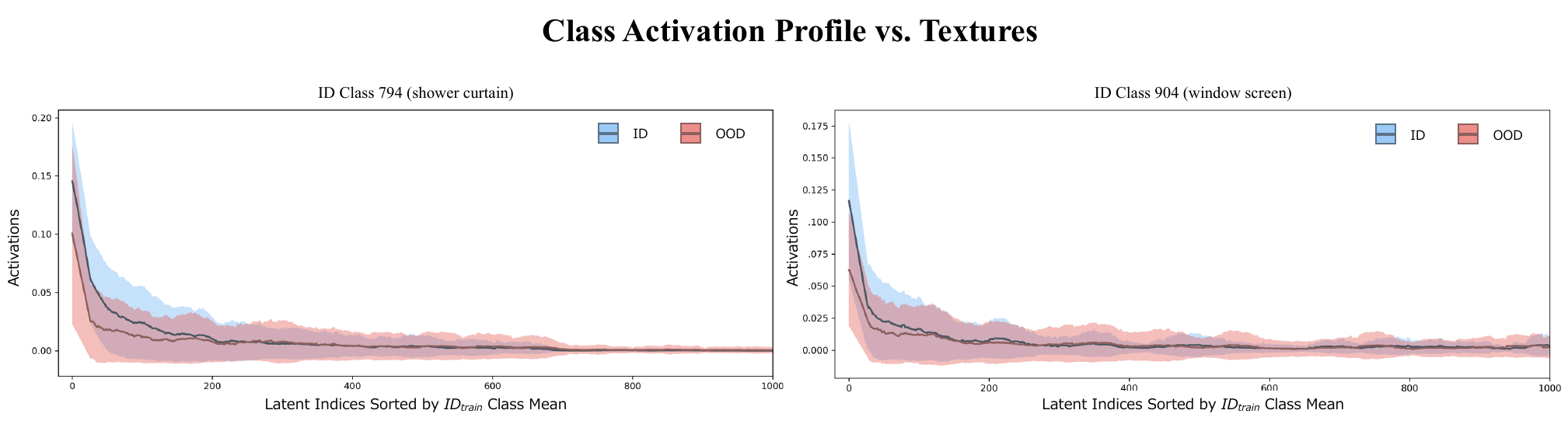}
        \label{fig:activation_appendix9-4}
    \end{subfigure}
    
    \caption{\textbf{Consistency of structural activation profiles across diverse OOD datasets.} This figure provides additional examples comparing the mean activation on CAP of ID samples (Blue) and OOD samples (Red) misclassified as the same class across four benchmark datasets: iNaturalist, NINCO, OpenImage-O, and Textures. The latent indices are sorted according to the mean ID activation profile of the respective class. In all cases, the distinctive activation head of ID samples is sharply attenuated and dispersed in OOD samples, confirming the generality of the structural disruption observed in the main paper.}
    \label{fig:structral_diff_appendix}
\end{figure}

To further substantiate the claims made in Section \ref{sec:profile} regarding the fundamental structural disruption between ID and misclassified OOD samples, Figure \ref{fig:structral_diff_appendix} presents comprehensive activation profile comparisons. This supplementary evidence confirms that the phenomena illustrated in Figure \ref{fig:IDvsOOD} are systematic findings, not exceptions limited to specific datasets or classes. As shown across all tested OOD benchmarks (iNaturalist, NINCO, OpenImage-O, and Textures), the consistent trend is the inability of OOD samples (Red) to replicate the activation signature of true ID data (Blue). Specifically, OOD activation profiles display a diffused and lower-intensity distribution, particularly failing to match the high-energy, concentrated peak—the ``head'' of the profile—that is characteristic of the ID samples. This consistent structural failure across varied OOD sources underscores the robustness of our central observation, and provides the essential foundation for our EPD metric.

\vspace{2.0em}
\section{Hyperparameter Ablation: Latent Dimension and Sparsity Level}
\label{sec:hyperparam_ablation}
 
\begin{figure}[h]
  \centering
   \includegraphics[width=1.0\linewidth]{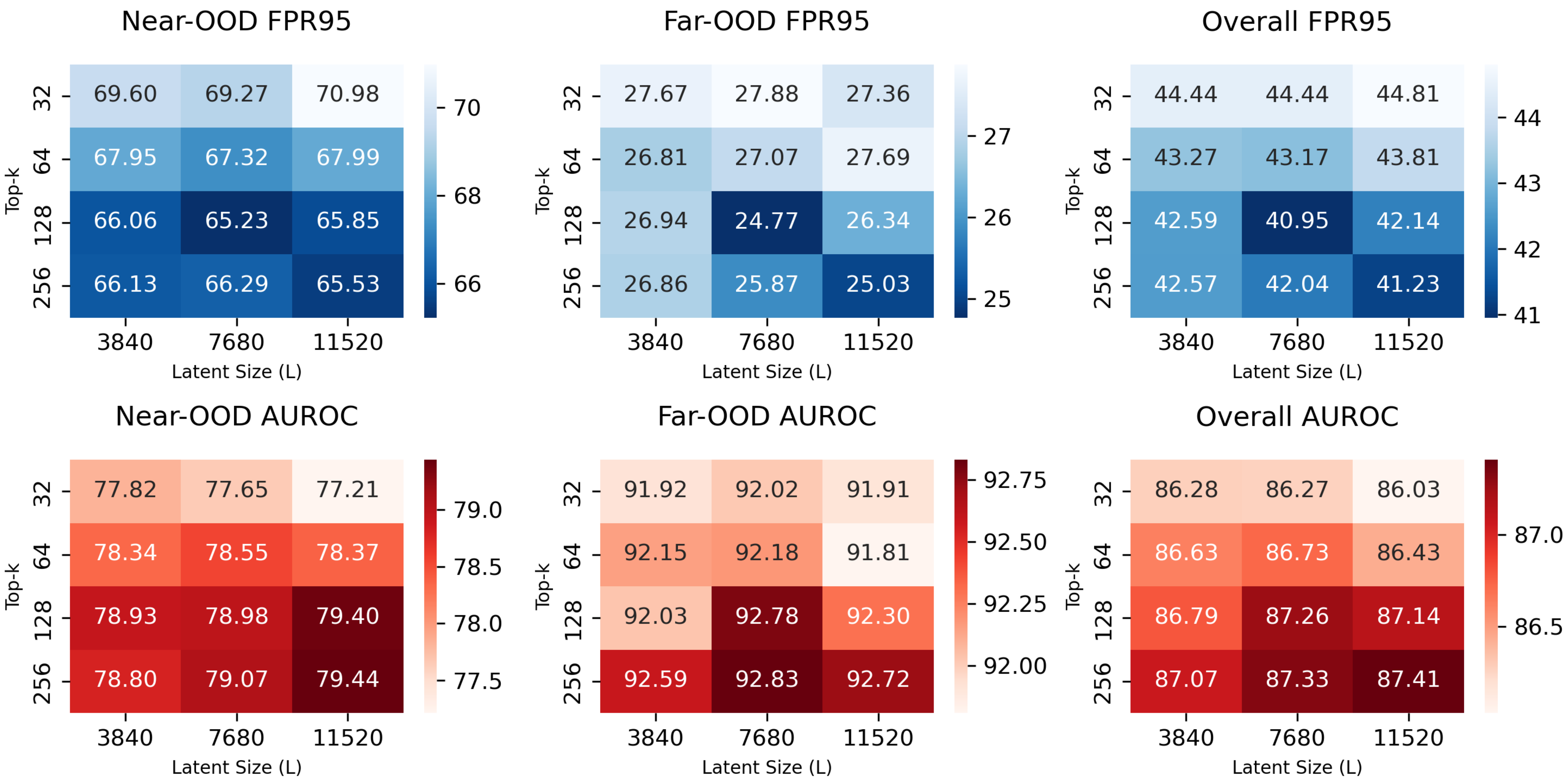}
   \caption{\textbf{Hyperparameter sensitivity analysis.} FPR95 (blue, lower is better) and AUROC (red, higher is better) across combinations of latent dimension ($L$) and sparsity level ($k$). Darker colors indicate better performance. The selected configuration ($L=7680$, $k=128$) minimizes overall FPR95 while maintaining strong AUROC, representing the optimal trade-off between robustness and separability.}
   \label{fig:perf_heatmap}
\end{figure}
 
Figure~\ref{fig:perf_heatmap} presents a comprehensive sensitivity analysis over our two primary architectural hyperparameters: the latent dimension $L \in \{3840, 7680, 11520\}$ and the sparsity level $k \in \{32, 64, 128, 256\}$. The results reveal a consistent pattern: increasing $L$ generally requires a proportional increase in $k$ to maintain reconstruction quality, but this scaling is eventually constrained by model instability at larger $k$ values. Our selected configuration ($L=7680$, $k=128$) yields the best overall FPR95 (40.96\%), prioritizing robustness in safety-critical detection over marginal AUROC gains. The relative flatness of AUROC across configurations further confirms that our method is not overly sensitive to the precise hyperparameter choice within a reasonable range.
 
\vspace{2.0em}
\section{Activation Head Size Sensitivity}

\begin{figure}[h]
  \centering
   \includegraphics[width=0.98\linewidth]{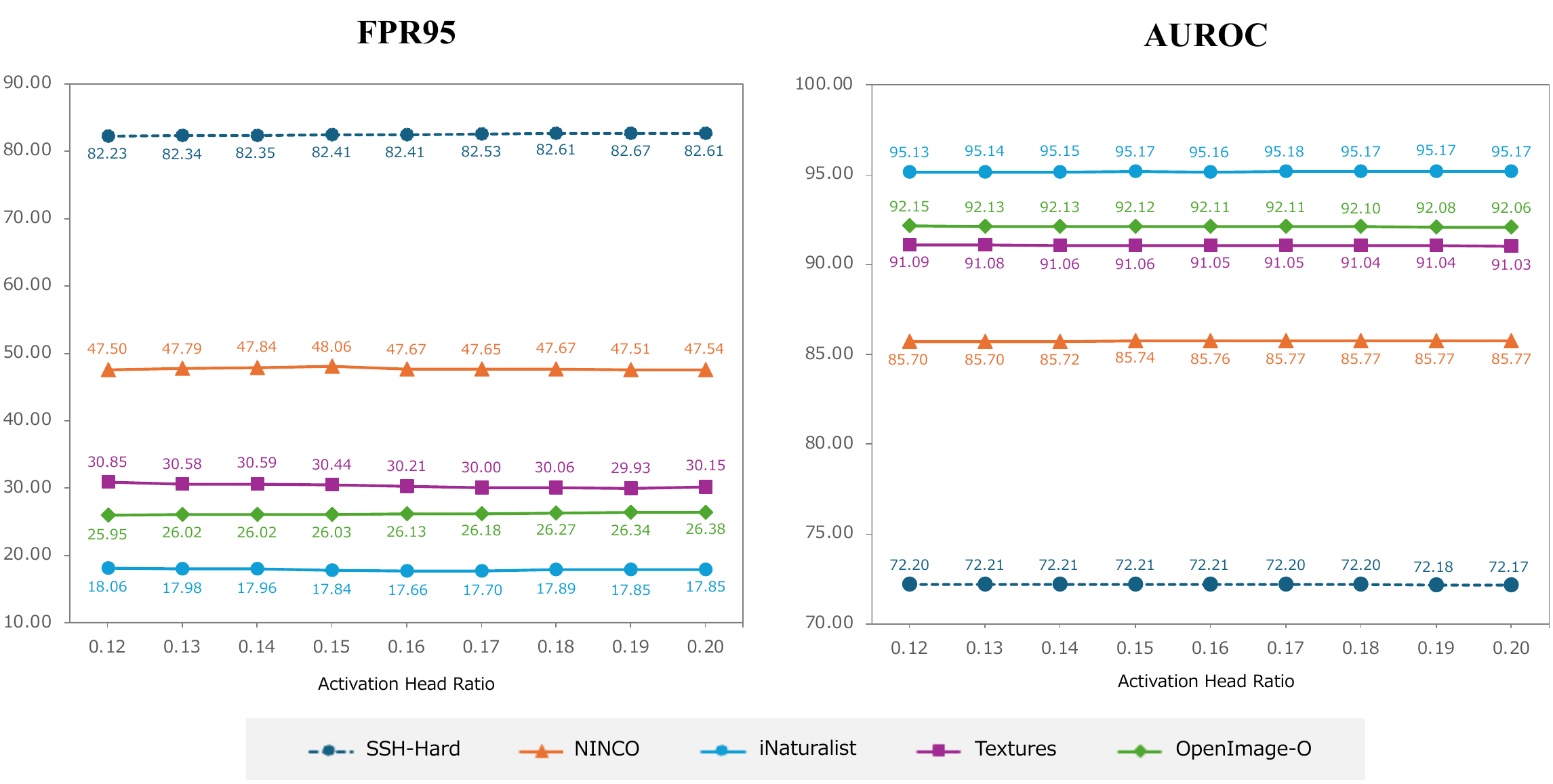}
    \caption{\textbf{Sensitivity analysis of the activation head ratio ($p$).} Performance metrics, FPR95 (Left) and AUROC (Right), are evaluated across various OOD benchmarks as a function of the activation head ratio ($p$). The ratio determines the size of the sorted latent feature set used for divergence calculation. The results demonstrate the stability of our proposed method EPD across the empirically derived meaningful range ($p=0.12$ to $p=0.20$), validating the robust selection of the chosen value ($p=0.15$) used in the main paper.}   
    \label{fig:act_head_appendix}
\end{figure}

To support the parameter choices detailed in Section \ref{sec:hyperparams}, Figure \ref{fig:act_head_appendix} provides a comprehensive sensitivity analysis on the Activation Head Ratio (p). This hyperparameter defines the percentage of top-activated latent indices utilized for calculating our EPD score.

The analysis systematically sweeps p across the empirically derived range. The resulting metrics, FPR95 and AUROC, show remarkable consistency across all tested benchmark OOD datasets (SSH-Hard, NINCO, iNaturalist, Textures, and OpenImage-O). The observed flatness of the performance curves explicitly confirms that the effectiveness of our EPD method is highly robust and not overly dependent on the precise selection of p. This validates our choice of p=0.15 as an optimal, stable setting that effectively balances detection sensitivity and generalizability across diverse OOD sources.

\section{Ablation: Scoring Metric}
\label{sec:metric_ablation_supp}
 
\begin{table}[h]
  \centering
  \resizebox{0.8\linewidth}{!}{
    \begin{tabular}{l c c c c c c} 
      \toprule
      \multirow{2}{*}{\textbf{Metric}} & 
      \multicolumn{2}{c}{\textbf{Near-OOD}} & 
      \multicolumn{2}{c}{\textbf{Far-OOD}} &
      \multicolumn{2}{c}{\textbf{Overall}} \\
      \cmidrule(lr){2-3} \cmidrule(lr){4-5} \cmidrule(lr){6-7}
      & \textbf{FPR95} & \textbf{AUROC} & \textbf{FPR95} & \textbf{AUROC} & \textbf{FPR95} & \textbf{AUROC} \\
      \midrule
      Euclidean   & 67.56 & 76.83 & 29.68 & 90.87 & 44.83 & 85.25  \\
      Cosine      & 71.74 & 77.09 & 47.93 & 84.21 & 57.45 & 81.36  \\
      \midrule
      \rowcolor{black!10}
      \textbf{EPD (Ours)} & \textbf{65.23} & \textbf{78.98} & \textbf{24.77} & \textbf{92.78} & \textbf{40.96} & \textbf{87.26} \\
      \bottomrule
    \end{tabular}
  }
  \caption{\textbf{Metric Ablation on ViT-B/16.} EPD outperforms Euclidean and cosine distance across all evaluation splits, validating the use of KL divergence over normalized energy profiles.}
  \label{tab:metric}
\end{table}
 
EPD's advantage stems from its use of KL divergence over $L_1$-normalized profiles, which captures the \textit{shape} of the energy distribution rather than its magnitude or direction alone. Euclidean distance performs competitively on Far-OOD but degrades on Near-OOD, while cosine similarity is notably weaker on Far-OOD—confirming that structural shape within the sparse simplex is the most discriminative signal for OOD detection.
 
\vspace{2.0em}

\section{Full Experimental Results}
\subsection{Full Results on ViT}
\label{sec:full_result_vit}

\begin{table}[h]
    \centering
    \resizebox{\textwidth}{!}{
        \begin{tabular}{l c c c c c c c c c c c c c c c c} 
            \toprule
            \multirow{3}{*}{\textbf{Method}} & 
            \multicolumn{6}{c}{\textbf{Near-OOD}} & 
            \multicolumn{8}{c}{\textbf{Far-OOD}} &
            \multicolumn{2}{c}{\textbf{Overall}} \\
            \cmidrule(lr){2-7} \cmidrule(lr){8-15} \cmidrule(lr){16-17}
            & \multicolumn{2}{c}{\textbf{SSB-Hard}} 
            & \multicolumn{2}{c}{\textbf{NINCO}} 
            & \multicolumn{2}{c}{\textbf{Average}}
            & \multicolumn{2}{c}{\textbf{iNaturalist}} 
            & \multicolumn{2}{c}{\textbf{Textures}} 
            & \multicolumn{2}{c}{\textbf{OpenImage-O}}
            & \multicolumn{2}{c}{\textbf{Average}}
            & \multicolumn{2}{c}{\textbf{Average}} \\
            \cmidrule(lr){2-3} \cmidrule(lr){4-5} \cmidrule(lr){6-7} \cmidrule(lr){8-9} \cmidrule(lr){10-11} \cmidrule(lr){12-13} \cmidrule(lr){14-15} \cmidrule(lr){16-17}
            & \textbf{FPR95} & \textbf{AUROC} & \textbf{FPR95} & \textbf{AUROC} & \textbf{FPR95} & \textbf{AUROC} & \textbf{FPR95} & \textbf{AUROC}
            & \textbf{FPR95} & \textbf{AUROC} & \textbf{FPR95} & \textbf{AUROC} & \textbf{FPR95} & \textbf{AUROC} & \textbf{FPR95} & \textbf{AUROC} \\
            \midrule
            ASH         & 93.50 & 53.89 & 95.40 & 52.52 & 94.45 & 53.20 & 97.02 & 50.63 & 98.49 & 48.53 & 94.80 & 55.52 & 96.77 & 51.56 & 95.84 & 52.22 \\
            DICE        & 89.77 & 59.05 & 81.09 & 71.67 & 85.43 & 65.36 & 47.92 & 82.51 & 54.79 & 82.21 & 52.57 & 82.23 & 51.76 & 82.32 & 65.23 & 75.53 \\
            EBO         & 92.25 & 58.80 & 94.16 & 66.02 & 93.21 & 62.41 & 83.58 & 79.30 & 83.65 & 81.17 & 88.79 & 76.48 & 85.34 & 78.98 & 88.49 & 72.35 \\
            GEN         & \textbf{82.24} & 70.09 & 59.31 & 82.51 & 70.78 & 76.30 & 22.94 & 93.54 & 38.31 & 90.23 & 35.43 & 90.27 & 32.23 & 91.35 & 47.65 & 85.33 \\
            GradNorm    & 93.62 & 42.96 & 95.81 & 64.40 & 94.72 & 53.68 & 91.16 & 42.42 & 92.25 & 44.99 & 94.52 & 37.82 & 92.64 & 41.74 & 93.47 & 46.52 \\
            KNN         & 86.22 & 65.97 & 54.73 & 82.25 & 70.48 & 74.11 & 27.74 & 91.46 & \underline{33.23} & \underline{91.12} & 34.82 & 89.86 & 31.93 & 90.81 & 47.35 & 84.13 \\
            MDS         & \underline{83.47} & \underline{71.57} & \underline{48.76} & \underline{86.52} & \underline{66.11} & \underline{79.04} & 20.66 & \underline{96.01} & 38.90 & 89.41 & 30.35 & \textbf{92.38} & 29.97 & \underline{92.60} & \underline{44.43} & \underline{87.18} \\
            MLS         & 91.52 & 64.20 & 92.98 & 72.40 & 92.25 & 68.30 & 72.98 & 85.29 & 78.93 & 83.74 & 85.78 & 81.60 & 79.23 & 83.54 & 84.44 & 77.45 \\
            MSP         & 86.41 & 68.94 & 77.35 & 78.11 & 81.88 & 73.53 & 42.42 & 88.19 & 56.44 & 85.06 & 56.11 & 84.87 & 51.66 & 86.04 & 63.75 & 81.03 \\
            OpenMax     & 89.20 & 68.60 & 88.36 & 78.68 & 88.78 & 73.64 & \underline{19.56} & 94.93 & 73.17 & 85.52 & 73.74 & 87.36 & 55.49 & 89.27 & 68.81 & 83.02 \\
            ReAct       & 90.46 & 63.10 & 78.50 & 75.43 & 84.48 & 69.27 & 48.22 & 86.11 & 55.87 & 86.66 & 57.68 & 84.29 & 53.92 & 85.69 & 66.15 & 79.12 \\
            RMDS        & 84.53 & \textbf{72.87} & \textbf{46.22} & \textbf{87.31} & \underline{65.38} & \textbf{80.09} & \underline{19.46} & \textbf{96.10} & 37.23 & 89.38 & \underline{29.57} & \underline{92.32} & \underline{28.75} & \underline{92.60} & \underline{43.40} & \textbf{87.60} \\
            SHE         & 85.74 & 68.04 & 56.01 & 84.18 & 70.88 & 76.11 & 22.17 & 93.57 & \textbf{25.65} & \textbf{92.65} & 33.59 & 91.04 & \underline{27.14} & 92.42 & 44.63 & 85.89 \\
            TempScale   & 87.36 & 68.55 & 81.90 & 77.80 & 84.63 & 73.18 & 43.08 & 88.54 & 58.22 & 85.39 & 60.00 & 85.04 & 53.77 & 86.32 & 66.11 & 81.06 \\
            ViM         & 90.06 & 69.42 & 57.45 & 84.64 & 73.76 & 77.03 & \textbf{17.59} & \underline{95.72} & 40.41 & 90.61 & \underline{29.59} & \underline{92.18} & 29.20 & \textbf{92.84} & 47.02 & 86.51 \\
            \midrule
            \rowcolor{black!10}
            \textbf{Ours} & \underline{82.41} & \underline{72.21} & \underline{48.06} & \underline{85.74} & \textbf{65.23} & \underline{78.98} & \underline{17.84} & 95.17 & \underline{30.44} & \underline{91.06} & \textbf{26.03} & 92.12 & \textbf{24.77} & \underline{92.78} & \textbf{40.96} & \underline{87.26} \\
            \bottomrule
        \end{tabular} 
    }
    \caption{OOD detection performance on ImageNet-1K benchmarks, evaluated on a ViT-B/16  (ID Acc: 81.14\%). For each metric, the best result is in \textbf{bold} and the second and third results are \underline{underlined}.}
    \label{tab:full_result-vit-reorganized}
\end{table}

\subsection{Full Results on Swin-T}

\begin{table}[h]
    \centering
    \resizebox{\textwidth}{!}{
        \begin{tabular}{l c c c c c c c c c c c c c c c c} 
            \toprule
            \multirow{3}{*}{\textbf{Method}} & 
            \multicolumn{6}{c}{\textbf{Near-OOD}} & 
            \multicolumn{8}{c}{\textbf{Far-OOD}} &
            \multicolumn{2}{c}{\textbf{Overall}} \\
            \cmidrule(lr){2-7} \cmidrule(lr){8-15} \cmidrule(lr){16-17}
            & \multicolumn{2}{c}{\textbf{SSB-Hard}} 
            & \multicolumn{2}{c}{\textbf{NINCO}} 
            & \multicolumn{2}{c}{\textbf{Average}}
            & \multicolumn{2}{c}{\textbf{iNaturalist}} 
            & \multicolumn{2}{c}{\textbf{Textures}} 
            & \multicolumn{2}{c}{\textbf{OpenImage-O}}
            & \multicolumn{2}{c}{\textbf{Average}}
            & \multicolumn{2}{c}{\textbf{Average}} \\
            \cmidrule(lr){2-3} \cmidrule(lr){4-5} \cmidrule(lr){6-7} \cmidrule(lr){8-9} \cmidrule(lr){10-11} \cmidrule(lr){12-13} \cmidrule(lr){14-15} \cmidrule(lr){16-17}
            & \textbf{FPR95} & \textbf{AUROC} & \textbf{FPR95} & \textbf{AUROC}
            & \textbf{FPR95} & \textbf{AUROC} & \textbf{FPR95} & \textbf{AUROC}
            & \textbf{FPR95} & \textbf{AUROC} & \textbf{FPR95} & \textbf{AUROC}
            & \textbf{FPR95} & \textbf{AUROC} & \textbf{FPR95} & \textbf{AUROC} \\
            \midrule
            ASH         & 95.22 & 46.28 & 93.98 & 47.00 & 94.60 & 46.64 & 94.19 & 46.49 & 96.17 & 41.32 & 93.99 & 45.22 & 94.78 & 44.34 & 94.71 & 45.26 \\
            DICE        & 95.76 & 49.97 & 97.28 & 50.00 & 96.52 & 49.99 & 98.11 & 47.09 & 87.37 & 77.61 & 96.36 & 58.67 & 93.95 & 61.12 & 94.98 & 56.67 \\
            EBO         & 87.51 & 68.28 & 79.23 & 78.42 & 83.37 & 73.35 & 61.18 & 85.17 & 84.29 & 79.00 & 80.20 & 80.24 & 75.22 & 81.47 & 78.48 & 78.22 \\
            GEN         & \textbf{79.35} & \textbf{72.78} & \textbf{48.10} & \textbf{85.16} & \textbf{63.73} & \textbf{78.97} & 20.47 & 94.23 & 45.04 & 88.16 & 32.77 & 90.60 & 32.76 & 91.00 & 45.15 & \underline{86.19} \\
            GradNorm    & 93.37 & 50.43 & 94.06 & 45.01 & 93.72 & 47.72 & 95.18 & 38.28 & 97.80 & 34.75 & 96.83 & 34.00 & 96.60 & 35.68 & 95.45 & 40.49 \\
            KNN         & 85.08 & 64.21 & 58.14 & 79.16 & 71.61 & 71.69 & 31.15 & 88.91 & \underline{35.79} & \underline{90.56} & 36.30 & 88.62 & 34.41 & 89.36 & 49.29 & 82.29 \\
            MDS         & 83.72 & 68.69 & 53.55 & 81.78 & 68.64 & 75.24 & 21.89 & 93.60 & 37.47 & 89.82 & 30.89 & 90.98 & 30.08 & 91.47 & 45.50 & 84.97 \\
            MLS         & 86.60 & 70.47 & 75.78 & 80.95 & 81.19 & 75.71 & 49.65 & 89.01 & 79.94 & 81.70 & 73.96 & 83.93 & 67.85 & 84.88 & 73.19 & 81.21 \\
            MSP         & \underline{81.02} & 71.75 & 60.36 & 81.69 & 70.69 & 76.72 & 37.50 & 89.84 & 61.49 & 83.27 & 48.97 & 85.81 & 49.32 & 86.31 & 57.87 & 82.47 \\
            OpenMax     & 85.54 & 71.52 & 72.85 & 81.69 & 79.20 & 76.61 & \underline{19.56} & \underline{95.06} & 76.41 & 82.81 & 60.88 & 87.75 & 52.28 & 88.54 & 63.05 & 83.77 \\
            ReAct       & 85.01 & 69.36 & 60.08 & 82.12 & 72.55 & 75.74 & 31.34 & 90.08 & 53.98 & 87.04 & 42.54 & 87.85 & 42.62 & 88.32 & 54.59 & 83.29 \\
            RMDS        & 82.66 & \underline{71.81} & \underline{49.89} & \underline{84.91} & \underline{66.28} & \underline{78.36} & \underline{18.55} & \textbf{95.57} & 39.09 & 89.26 & \underline{29.34} & \underline{92.10} & \underline{29.00} & \underline{92.31} & \underline{43.91} & \textbf{86.73} \\
            SHE         & 85.03 & 70.75 & 67.53 & 82.74 & 76.28 & 76.75 & 33.25 & 92.78 & 51.32 & 88.86 & 52.88 & 88.04 & 45.82 & 89.89 & 58.00 & 84.63 \\
            TempScale   & 82.12 & \underline{71.83} & 63.02 & 82.09 & 72.57 & \underline{76.96} & 36.84 & 90.36 & 62.89 & 83.68 & 50.30 & 86.20 & 50.01 & 86.75 & 59.03 & 82.83 \\
            ViM         & 88.55 & 68.94 & 60.80 & 81.85 & 74.68 & 75.40 & \textbf{17.98} & \underline{94.62} & \textbf{29.46} & \textbf{92.69} & \textbf{26.62} & \textbf{92.29} & \textbf{24.69} & \textbf{93.20} & \underline{44.68} & \underline{86.08} \\
            \midrule
            \rowcolor{black!10}
            \textbf{Ours} & \underline{81.62} & 70.50 & 55.40 & \underline{83.01} & \underline{68.51} & 76.76 & 20.37 & 94.07 & \underline{32.64} & \underline{90.86} & \underline{27.90} & \underline{91.71} & \underline{26.97} & \underline{92.21} & \textbf{43.59} & 86.03 \\
            \bottomrule
        \end{tabular}
    }
    \caption{OOD detection performance on ImageNet-1K benchmarks, evaluated on a Swin Transformer (ID Acc: 81.60\%. For each metric, the best result is in \textbf{bold} and the second and third results are \underline{underlined}.}
    \label{tab:full_result-swint-reorganized}
\end{table}

\subsection{Full Results on Multiple Architectures (Ours Only)}
\begin{table}[h]
    \centering
    \resizebox{\textwidth}{!}{
        \begin{tabular}{l c c c c c c c c c c c c c c c c} 
            \toprule
            \multirow{3}{*}{\textbf{Method}} & 
            \multicolumn{6}{c}{\textbf{Near-OOD}} & 
            \multicolumn{8}{c}{\textbf{Far-OOD}} &
            \multicolumn{2}{c}{\textbf{Overall}} \\
            \cmidrule(lr){2-7} \cmidrule(lr){8-15} \cmidrule(lr){16-17}
            & \multicolumn{2}{c}{\textbf{SSB-Hard}} 
            & \multicolumn{2}{c}{\textbf{NINCO}} 
            & \multicolumn{2}{c}{\textbf{Average}}
            & \multicolumn{2}{c}{\textbf{iNaturalist}} 
            & \multicolumn{2}{c}{\textbf{Textures}} 
            & \multicolumn{2}{c}{\textbf{OpenImage-O}}
            & \multicolumn{2}{c}{\textbf{Average}}
            & \multicolumn{2}{c}{\textbf{Average}} \\
            \cmidrule(lr){2-3} \cmidrule(lr){4-5} \cmidrule(lr){6-7} \cmidrule(lr){8-9} \cmidrule(lr){10-11} \cmidrule(lr){12-13} \cmidrule(lr){14-15} \cmidrule(lr){16-17}
            & \textbf{FPR95} & \textbf{AUROC} & \textbf{FPR95} & \textbf{AUROC}
            & \textbf{FPR95} & \textbf{AUROC} & \textbf{FPR95} & \textbf{AUROC}
            & \textbf{FPR95} & \textbf{AUROC} & \textbf{FPR95} & \textbf{AUROC}
            & \textbf{FPR95} & \textbf{AUROC} & \textbf{FPR95} & \textbf{AUROC} \\
            \midrule
            \textbf{Ours (ViT-B)} & 82.41 & 72.21 & 48.06 & 85.74 & 65.23 & 78.98 & 17.84 & 95.17 & 30.44 & 91.06 & 26.03 & 92.12 & 24.77 & 92.78 & 40.96 & 87.26 \\
            \textbf{Ours (Swin-T)} & 81.62 & 70.50 & 55.40 & 83.01 & 68.51 & 76.76 & 20.37 & 94.07 & 32.64 & 90.86 & 27.90 & 91.71 & 26.97 & 92.21 & 43.59 & 86.03 \\
            \textbf{Ours (DINOv2)} & 80.05 & 76.82 & 44.74 & 89.63 & 62.40 & 83.23 & 7.21 & 97.87 & 32.35 & 92.11 & 16.22 & 95.51 & 18.59 & 95.16 & 40.50 & 89.20 \\
            \bottomrule
        \end{tabular}
    }
    \caption{\textbf{OOD Detection performance of our method across different architectures.} Detailed OOD detection performance on ImageNet-1K benchmarks for our proposed method (EPD) applied to DINOv2 B/14 (ID Acc: 84.64\%), ViT-B/16 (ID Acc: 81.14\%), and Swin-T (ID Acc: 81.60\%).}
    \label{tab:full_result-ours-arch-comparison}
\end{table}

The comprehensive OOD detection performance results for the ViT-B/16 (Table \ref{tab:full_result-vit-reorganized}) and Swin-T (Table \ref{tab:full_result-swint-reorganized}) backbones, including detailed metrics for all OpenOOD benchmark methods, are presented here. Moreover, to definitively demonstrate the generalizability and robust effectiveness of our core structural detection mechanism, Table \ref{tab:full_result-ours-arch-comparison} provides a side-by-side comparison of our method applied to three distinct Vision Transformer backbones: DINOv2, ViT-B/16, and Swin-T. These tables provide the detailed data that underpins the summary findings discussed in Section \ref{sec:vit_results} and \ref{sec:other_backbones}.

\begin{itemize}
    \item \textbf{ViT-B/16}: The detailed results emphatically confirm that our method, EPD using CAP, achieves the best overall average FPR95 ($\mathbf{40.96\%}$) among all compared methods, outperforming the next best methods (RMDS at $43.40\%$ and MDS at $44.43\%$). This performance gap validates EPD's exceptional capability in minimizing false positives across diverse OOD scenarios, which is paramount for safety-critical applications.
    
    \item \textbf{Swin-T}: Our method achieves the best overall average FPR95 ($\mathbf{43.59\%}$) among the compared methods on this architecture. The overall AUROC reflects the influence of the Swin-T architecture, where localized attention and hierarchical feature extraction result in latent representations with reduced global coherence. Despite this, our method secures the best overall average FPR95 and a competitive AUROC score.
    
    \item \textbf{DINOv2}: Our method achieves its highest overall performance on DINOv2, leading to the best overall FPR95 ($\mathbf{40.50\%}$) among all tested backbones. This inclusion of DINOv2 further validates the general applicability of EPD, proving that our method's efficacy of OOD detection stems from fundamental structural properties of the latent space. It is important to note that due to the lack of comprehensive OOD benchmark results for DINOv2 in existing literature, Table \ref{tab:full_result-ours-arch-comparison} presents the performance of only our method as an internal comparison.
\end{itemize}

This comparative analysis solidifies our hypothesis that the efficacy of OOD detection through structural deviation is not dependent on a specific backbone model.

\section{Results on Different ID Dataset: CIFAR-100}

\begin{table}[h]
    \centering
    \resizebox{\textwidth}{!}{
        \begin{tabular}{l c c c c c c c c c c c c c c c c} 
            \toprule
            \multirow{3}{*}{\textbf{Method}} & 
            \multicolumn{6}{c}{\textbf{Near-OOD}} & 
            \multicolumn{8}{c}{\textbf{Far-OOD}} &
            \multicolumn{2}{c}{\textbf{Overall}} \\
            \cmidrule(lr){2-7} \cmidrule(lr){8-15} \cmidrule(lr){16-17}
            & \multicolumn{2}{c}{\textbf{CIFAR10}} 
            & \multicolumn{2}{c}{\textbf{TinyImageNet}} 
            & \multicolumn{2}{c}{\textbf{Average}}
            & \multicolumn{2}{c}{\textbf{SVHN}} 
            & \multicolumn{2}{c}{\textbf{Textures}}
            & \multicolumn{2}{c}{\textbf{Places365}}
            & \multicolumn{2}{c}{\textbf{Average}}
            & \multicolumn{2}{c}{\textbf{Average}} \\
            \cmidrule(lr){2-3} \cmidrule(lr){4-5} \cmidrule(lr){6-7} \cmidrule(lr){8-9} \cmidrule(lr){10-11} \cmidrule(lr){12-13} \cmidrule(lr){14-15} \cmidrule(lr){16-17}
            & \textbf{FPR95} & \textbf{AUROC} & \textbf{FPR95} & \textbf{AUROC}
            & \textbf{FPR95} & \textbf{AUROC} & \textbf{FPR95} & \textbf{AUROC}
            & \textbf{FPR95} & \textbf{AUROC} & \textbf{FPR95} & \textbf{AUROC}
            & \textbf{FPR95} & \textbf{AUROC} & \textbf{FPR95} & \textbf{AUROC}  \\
            \midrule
            ASH         & 87.04 & 56.96 & 90.97 & 51.23 & 89.01 & 54.09 & 81.59 & 57.61 & 82.07 & 58.53 & 91.72 & 52.57 & 85.13 & 56.24 & 86.68 & 55.38 \\
            DICE        & 69.63	& 81.68 & 60.53 & 83.57 & 65.08 & 82.63 & 44.10 & \underline{90.96} & 73.32 & 82.57 & 65.11 & 82.65 & 60.84 & 85.39 & 62.54 & 84.29 \\
            EBO         & 63.84 & 84.38 & 49.64 & 86.83 & 56.74 & 85.60 & \underline{39.33} & 89.89 & 59.26 & 86.88 & 64.07 & 83.81 & 54.22 & 86.86 & 55.23 & 86.36 \\
            GEN         & 62.46 & 84.64 & 49.98 & 86.80 & 56.22 & 85.72 & 40.06 & 89.63 & 58.78 & 86.89 & 64.43 & 83.80 & 54.42 & 86.77 & 55.14 & 86.35 \\
            GradNorm    & 48.78 & 85.52 & 45.17 & 87.63 & 46.97 & 86.57 & 49.54 & \textbf{91.47} & 58.10 & 86.65 & \textbf{44.07} & \textbf{88.41} & 50.57 & \textbf{88.84} & 49.13 & \underline{87.94} \\
            KNN         & \textbf{43.61} & \underline{87.67} & \underline{41.53} & \underline{87.77} & \underline{42.57} & \underline{87.72} & 41.08 & 88.95 & 47.88 & \underline{87.96} & 50.66 & 85.39 & 46.54 & 87.43 & \underline{44.95} & \underline{87.55} \\
            MDS         & \underline{44.64}	& \underline{88.10}	& 44.81	& 85.60	& 44.73	& 86.85	& 50.50	& 78.14	& \underline{38.52}	& \underline{90.14}	& 56.47	& 82.55	& 48.50 & 83.61 & 46.99 & 84.91 \\
            MLS         & 63.70	& 84.17	& 49.52	& 86.62	& 56.61	& 85.39	& 39.57	& 89.46	& 59.27	& 86.51	& 64.07	& 83.58	& 54.30 & 86.52 & 55.23 & 86.07 \\
            MSP         & 61.80	& 81.68	& 52.63	& 83.84	& 57.22	& 82.76	& 46.98	& 84.81	& 64.86	& 82.25	& 64.56	& 80.80	& 58.80 & 82.62 & 58.17 & 82.68 \\
            OpenMax     & 62.94 & 80.95 & 54.92 & 83.58 & 58.93 & 82.26 & 48.94 & 84.42 & 75.71 & 79.82 & 66.20 & 81.15 & 63.62 & 81.80 & 61.74 & 81.98 \\
            ReAct       & 51.21 & 86.02 & \underline{41.68} & \underline{88.11} & 46.44 & 87.06 & 41.10 & \underline{91.04} & 57.11 & 87.31 & 47.42 & \underline{87.14} & 48.54 & \underline{88.50} & 47.70 & 87.92 \\
            RMDS        & 47.02 & 86.50 & 41.81 & 86.82 & 44.42 & 86.66 & \underline{37.60} & 87.27 & \underline{43.03} & 86.84 & 49.79 & 84.64 & \underline{43.47} & 86.25 & \underline{43.85} & 86.41 \\
            SHE         & 49.91 & 86.03 & 46.61 & 87.35 & 48.26 & 86.69 & 78.93 & 88.52 & 65.28 & 86.85 & \underline{46.96} & \underline{87.40} & 63.72 & 87.59 & 57.54 & 87.23 \\
            TempScale   & 62.41 & 83.02 & 50.92 & 85.18 & 56.67 & 84.10 & 44.23 & 86.85 & 61.70 & 84.23 & 64.19 & 82.13 & 56.71 & 84.40 & 56.69 & 84.28 \\
            ViM         & \underline{43.89} & \textbf{88.58} & \textbf{38.73} & \textbf{88.43} & \textbf{41.31} & \textbf{88.50} & \textbf{36.11} & 86.88 & \textbf{35.86} & \textbf{91.91} & \underline{46.51} & 86.53 & \textbf{39.49} & \underline{88.44} & \textbf{40.22} & \textbf{88.47} \\
            \midrule
            \rowcolor{black!10}
            \textbf{Ours} & 44.76 & 87.25 & 42.76 & 87.48 & \underline{43.76} & \underline{87.37} & 41.76 & 87.92 & 45.21 & 87.93 & 52.44 & 85.21 & \underline{46.47} & 87.02 & 45.39 & 87.16 \\
            \bottomrule
        \end{tabular}
    }
    \caption{OOD detection performance on CIFAR-100 benchmarks, evaluated on a ViT-B/16 (ID Acc: 84.78\%). For this experiment, we employed a fine-tuned version of \texttt{google/vit-base-patch16-224-in21k}. For each metric, the best result is in \textbf{bold} and the second and third results are \underline{underlined}.}
    \label{tab:full_result_vit_cifar100}
\end{table}

To validate the general applicability and domain robustness of our EPD method, we conducted extensive experiments using CIFAR-100 as the ID dataset, a domain fundamentally different in scale and complexity from the ImageNet domain studied in the main experiment. The full set of results comparing EPD with established OpenOOD benchmarks is detailed in Table \ref{tab:full_result_vit_cifar100}.
The OOD datasets utilized for this analysis were selected from the suggested list for CIFAR-100 in the OpenOOD v 1.5 framework.
The results demonstrate that EPD maintains its effectiveness despite the significant shift in the training domain. While some methods, such as KNN and GEN, showing a notable surge in performance on the CIFAR-100 domain compared to their ImageNet-1k score, our method demonstrates cross-domain stability. EPD maintains a robust performance where the score gap relative to the top-ranking method is small. Our method shows highly stable performance across both Near-OOD and Far-OOD categories, proving that the core mechanism is not architecture- or domain-specific. 

\clearpage
 
\section{Comparison with Recent Baselines: MDS++ and RMDS++}
\label{sec:recent_baselines}
 
Table~\ref{tab:comparison-recent} extends our comparison to include recently proposed methods MDS++ and RMDS++~\cite{mueller2025mahalanobis++}, which are enhanced variants of MDS and RMDS. Note that unlike the other baselines evaluated via OpenOOD v1.5, these results were obtained through our own implementation. Our method remains highly competitive against these stronger baselines. In particular, our approach records the best FPR95 on SSB-Hard and Textures among all three methods, demonstrating that our structural detection mechanism provides complementary advantages. While MDS++ and RMDS++ outperform ours on iNaturalist and NINCO, our method shows superior or competitive performance on the remaining datasets.
 
\begin{table}[h]
    \centering
    \resizebox{\linewidth}{!}{
        \begin{tabular}{l c c c c c c c c c c} 
            \toprule
            \multirow{3}{*}{\textbf{Method}} & 
            \multicolumn{4}{c}{\textbf{Near-OOD}} & 
            \multicolumn{6}{c}{\textbf{Far-OOD}} \\
            \cmidrule(lr){2-5} \cmidrule(lr){6-11}
            & \multicolumn{2}{c}{\textbf{SSB-Hard}} 
            & \multicolumn{2}{c}{\textbf{NINCO}} 
            & \multicolumn{2}{c}{\textbf{iNaturalist}} 
            & \multicolumn{2}{c}{\textbf{Textures}} 
            & \multicolumn{2}{c}{\textbf{OpenImage-O}} \\
            \cmidrule(lr){2-3} \cmidrule(lr){4-5} \cmidrule(lr){6-7} \cmidrule(lr){8-9} \cmidrule(lr){10-11} 
            & \textbf{FPR95} & \textbf{AUROC} & \textbf{FPR95} & \textbf{AUROC} & \textbf{FPR95} & \textbf{AUROC} & \textbf{FPR95} & \textbf{AUROC}
            & \textbf{FPR95} & \textbf{AUROC} \\
            \midrule
            MDS++         & 83.97 & \underline{74.04} & \underline{44.90} & \underline{88.55} & \textbf{11.97} & \textbf{97.18} & \underline{31.37} & \underline{90.57} & \textbf{25.21} & \textbf{93.11} \\
            RMDS++        & \underline{83.29} & \textbf{74.81} & \textbf{43.86} & \textbf{88.67} & \underline{13.34} & \underline{96.95} & 34.74 & 89.28 & 26.45 & \underline{92.66} \\
            \midrule
            \rowcolor{black!10}
            \textbf{Ours} & \textbf{82.41} & 72.21 & 48.06 & 85.74 & 17.84 & 95.17 & \textbf{30.44} & \textbf{91.06} & \underline{26.03} & 92.12 \\
            \bottomrule
        \end{tabular} 
    }
    \caption{Comparison with recent methods on ImageNet-1K benchmarks using ViT-B/16. \textbf{Bold}: best result; \underline{underlined}: second best.}
    \label{tab:comparison-recent}
\end{table}
 
\section{CAP Cosine Similarity Analysis}
\label{sec:cap_analysis}
 
\begin{figure}[h]
  \centering
  \includegraphics[width=1.0\linewidth]{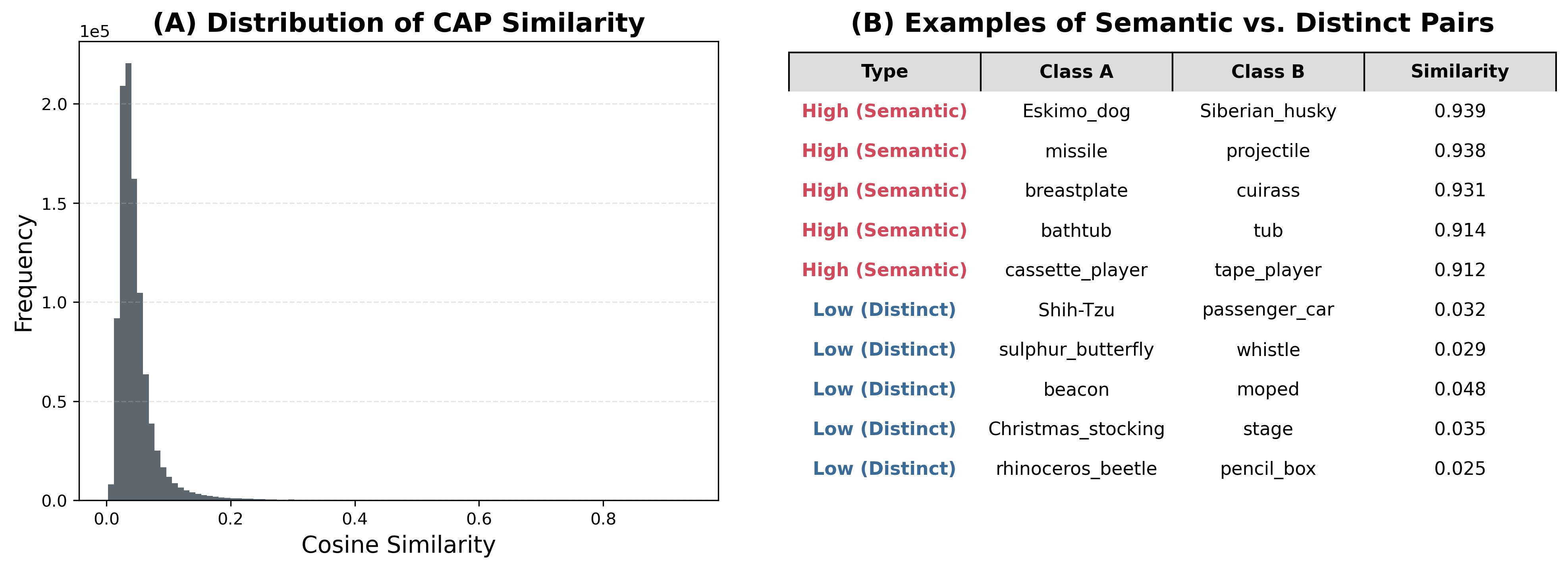}
  \caption{\textbf{CAP cosine similarity analysis.} \textbf{(A)} Distribution of pairwise cosine similarity between all 1,000 ID class CAPs. The distribution peaks near zero, confirming that most ID classes form mutually orthogonal subspaces in the sparse latent space. \textbf{(B)} Examples of class pairs with high and low CAP similarity. High-similarity pairs (red) correspond to semantically related concepts (\eg, Eskimo dog vs.\ Siberian husky), while low-similarity pairs (blue) represent semantically distinct classes. This validates that CAPs capture meaningful semantic identity rather than arbitrary activation patterns.}
  \label{fig:cap_similarity}
\end{figure}
 
The pairwise cosine similarity distribution in Figure~\ref{fig:cap_similarity}(A) provides further evidence for the disjointness of CAPs established in Section~\ref{sec:3.1} of the main paper. While the Jaccard similarity in Figure~2 measures binary feature overlap, cosine similarity here captures the directional alignment of the full mean activation vectors. The zero-skewed distribution confirms that the vast majority of ID class pairs occupy nearly orthogonal subspaces in the sparse latent space. The qualitative examples in (B) further validate that the few high-similarity pairs correspond exclusively to semantically related classes (\eg, near-synonym animal breeds or synonymous objects), demonstrating that the sparse latent space organizes classes according to genuine semantic relationships rather than arbitrary activation patterns.

\end{document}